\documentclass[10pt,twocolumn,letterpaper]{article}

\usepackage{cvpr}              

\usepackage{amsmath,amsfonts,bm}

\def\eqref#1{equation~\ref{#1}}

\def\1{\bm{1}}

\def\vs{{\bm{s}}}

\DeclareMathAlphabet{\mathsfit}{\encodingdefault}{\sfdefault}{m}{sl}
\SetMathAlphabet{\mathsfit}{bold}{\encodingdefault}{\sfdefault}{bx}{n}

\usepackage{url}            
\usepackage{booktabs}       
\usepackage{amsfonts}       
\usepackage{nicefrac}       
\usepackage{microtype}      
\usepackage{acronym}
\usepackage{marvosym}
\usepackage{wasysym} 
\usepackage[most,skins]{tcolorbox}
\usepackage{xcolor}

\newtcolorbox{promptbox}{
  enhanced,
  breakable,
  colback=gray!5,
  colframe=gray!100,
  boxrule=0.5pt,
  sharp corners,
  top=0.5ex, bottom=0.5ex,  
  leftrule=0.5pt, rightrule=0.5pt,
  toprule at break=0mm,     
  bottomrule at break=0mm   
}

\usepackage{caption}
\usepackage{makecell}
\usepackage{graphicx}
\usepackage{xspace}
\usepackage{amsmath}
\usepackage{adjustbox}
\usepackage{multirow}
\usepackage{pifont}
\usepackage{bm}
\usepackage{svg}
\usepackage{subfiles}
\usepackage{wasysym}
\usepackage{wrapfig}
\usepackage{threeparttable}
\usepackage{tabularx}
\usepackage{cuted}
\usepackage{caption}
\usepackage{float}

\definecolor{cvprblue}{rgb}{0.21,0.49,0.74}
\usepackage[pagebackref,breaklinks,colorlinks,allcolors=cvprblue]{hyperref}

\def\paperID{*****} 
\def\confName{CVPR}
\def\confYear{2026}

\title{RichControl: Structure- and Appearance-Rich Training-Free Spatial Control for Text-to-Image Generation}

\author{
Lexi Pang\textsuperscript{1} \quad Liheng Zhang\textsuperscript{1} \quad Hang Ye\textsuperscript{1} \quad Xiaoxuan Ma\textsuperscript{2} \quad Yizhou Wang\textsuperscript{1} \\
\textsuperscript{1}Peking University \quad
\textsuperscript{2}Carnegie Mellon University \\
{\tt\small \{panglexi, zhangliheng\}@stu.pku.edu.cn, \{yehang, yizhou.wang\}@pku.edu.cn \ xiaoxuam@andrew.cmu.edu}  \\
}

\begin{document}

\newcommand{\yehang}[1]{\textcolor{orange}{\textbf{\small [#1 --yehang]}}}
\newcommand{\xiaoxuan}[1]{\textcolor{purple}{\textbf{\small [#1 --xiaoxuan]}}}
\newcommand{\liheng}[1]{\textcolor{purple}{\textbf{\small [#1 --liheng]}}}
\newcommand{\lexi}[1]{\textcolor{blue}{\textbf{\small [#1 --lexi]}}}
\makeatletter
\newcommand{\thickhline}{%
    \noalign {\ifnum 0=`}\fi \hrule height 1pt
    \futurelet \reserved@a \@xhline
}

\definecolor{mygray}{gray}{.95}
\definecolor{mygraycell}{gray}{.6}
\definecolor{mygreen}{HTML}{37C291}
\definecolor{myorange}{HTML}{FF8A3C}

\newcommand{\graycell}[1]{\textcolor{mygraycell}{#1}}

\newcommand{\cmark}{\ding{51}} 
\newcommand{\xmark}{\ding{55}}

\newcommand{\misscite}{\textcolor{red}{[C]~}}
\newcommand{\misscap}{\textcolor{red}{[Cap]~}}

\newcommand{\tobeshown}{\textcolor{red}{[S]~}}

\newcommand{\projpage}{\href{https://zhang-liheng.github.io/rich-control/}{https://zhang-liheng.github.io/rich-control/}\xspace}

\crefname{algorithm}{Alg.}{Algs.}
\Crefname{algocf}{Algorithm}{Algorithms}
\crefname{section}{Sec.}{Secs.}
\Crefname{section}{Section}{Sections}
\crefname{table}{Tab.}{Tabs.}
\Crefname{table}{Table}{Tables}
\crefname{figure}{Fig.}{Figs.}
\Crefname{figure}{Figure}{Figures}
\crefname{equation}{Eq.}{Eqs.}
\Crefname{equation}{Equation}{Equations}
\crefname{appendix}{Appx.}{Appxs.}
\Crefname{appendix}{Appendix}{Appendices}

\newcommand{\dname}{our method }
\newcommand{\appcontrol}{appearance-rich prompting }
\newcommand{\restart}{restart}



\newcommand{\timestep}{t}

\newcommand{\image}{\mathbf{I}}
\newcommand{\imagegen}{\image}
\newcommand{\imagecond}{\image^\text{struct}}
\newcommand{\imagetcond}{\image_{t}^\text{struct}}
\newcommand{\imageapp}{\image^\text{app}}
\newcommand{\imagenatural}{\image^\text{natural}}

\newcommand{\prompt}{\mathcal{P}}
\newcommand{\promptapp}{\prompt^\text{app}}

\newcommand{\injlast}{\tau}
\newcommand{\asyncfunc}{g(t)}


\newcommand{\width}{w}
\newcommand{\minwidth}{w_\text{min}}
\newcommand{\maxwidth}{w_\text{max}}
\newcommand{\kernelsize}{k^e}

\newcommand{\restarttimes}{N}
\newcommand{\restartsteps}{S}
\newcommand{\restarttmin}{t_\text{min}}
\newcommand{\restarttmax}{t_\text{max}}
\newcommand{\restartstdmin}{\sigma_{t_{\min}}}
\newcommand{\restartstdmax}{\sigma_{t_{\max}}}

\newcommand{\recurtimes}{N'}
\newcommand{\recurtmin}{t_\text{min}'}
\newcommand{\recurtmax}{t_\text{max}'}

\newcommand{\latent}{\mathbf{x}}
\newcommand{\latentzero}{\latent_{0}}
\newcommand{\latentt}{\latent_{t}}
\newcommand{\latenttminus}{\latent_{t-1}}
\newcommand{\latentgt}{\latent_{g(t)}}
\newcommand{\latentT}{\latent_{T}}

\newcommand{\latentzerogen}{\latentzero}
\newcommand{\latenttgen}{\latentt}
\newcommand{\latentgtgen}{\latentgt}
\newcommand{\latentTgen}{\latentT}

\newcommand{\latentzerocond}{\latentzero^\text{struct}}
\newcommand{\latenttcond}{\latentt^\text{struct}}
\newcommand{\latentgtcond}{\latentgt^\text{struct}}

\newcommand{\latentzeroapp}{\latentzero^\text{app}}
\newcommand{\latenttapp}{\latentt^\text{app}}
\newcommand{\latentgtapp}{\latentgt^\text{app}}
\newcommand{\latentTapp}{\latentT^\text{app}}

\newcommand{\latenttmin}{\latent_{t_\text{min}}}
\newcommand{\latenttminith}{\latenttmin^{(i)}}
\newcommand{\latenttzeroith}{\latenttmin^{(0)}}
\newcommand{\latenttminiplusoneth}{\latenttmin^{(i+1)}}
\newcommand{\latenttmax}{\latent_{t_\text{max}}}
\newcommand{\latenttmaxith}{\latenttmax^{(i)}}
\newcommand{\latenttmaxiplusoneth}{\latenttmax^{(i+1)}}
\newcommand{\noise}{\mathbf{\epsilon}}
\newcommand{\noisetmintotmax}{\noise_{t_\text{min}\to t_\text{max}}}
\newcommand{\latentpred}{\hat{\latent}}
\newcommand{\latentpredt}{\latentpred_{t}}
\newcommand{\latentpredtcond}{\latentpredt^\text{struct}}
\newcommand{\imagergb}{\image^\text{natural}}
\newcommand{\latentpredtrgb}{\latentpredt^\text{natural}}
\newcommand{\latentpredtmin}{\latentpred_{t_\text{min}}}
\newcommand{\latentpredtminzeroth}{\latentpredtmin^{(0)}}
\newcommand{\latentpredtminNminusoneth}{\latentpredtmin^{(N-1)}}

\newcommand{\feature}{\mathbf{f}}
\newcommand{\featurelt}{\feature_{l,t}}
\newcommand{\featurelgt}{\feature_{l,g(t)}}
\newcommand{\featureltcond}{\featurelt^\text{struct}}
\newcommand{\featurelgtcond}{\featurelgt^\text{struct}}
\newcommand{\featureltgen}{\featurelt}
\newcommand{\featureltrgb}{\featurelt^{\text{natural}}}
\newcommand{\attn}{\mathbf{A}}
\newcommand{\attnlt}{\attn_{l,t}}
\newcommand{\attnlgt}{\attn_{l,g(t)}}
\newcommand{\attnltcond}{\attnlt^\text{struct}}
\newcommand{\attnlgtcond}{\attnlgt^\text{struct}}
\newcommand{\attnltgen}{\attnlt}

\newcommand{\appctrl}{appearance-rich prompting}

\makeatletter
\DeclareRobustCommand\onedot{\futurelet\@let@token\@onedot}
\def\@onedot{\ifx\@let@token.\else.\null\fi\xspace}
\def\eg{\emph{e.g}\onedot} 
\def\Eg{\emph{E.g}\onedot}
\def\ie{\emph{i.e}\onedot} 
\def\Ie{\emph{I.e}\onedot}
\def\cf{\emph{c.f}\onedot} 
\def\Cf{\emph{C.f}\onedot}
\def\etc{\emph{etc}\onedot} 
\def\vs{\emph{vs}\onedot}
\def\aka{a.k.a\onedot}
\def\wrt{w.r.t\onedot} 
\def\dof{d.o.f\onedot}
\def\etal{\emph{et al}\onedot}
\makeatother
\newcommand{\tmin}{t_{\textrm{min}}}
\newcommand{\tmax}{t_{\textrm{max}}}

\acrodef{sota}[SOTA]{State-of-the-Art}


\newcommand{\loss}{\mathcal{L}}

\newcommand{\revision}[1]{\textcolor{red}{{#1}}}

\twocolumn[{%
    \renewcommand\twocolumn[1][]{#1}%
    \maketitle
    \begin{center}
        \includegraphics[width=\linewidth]{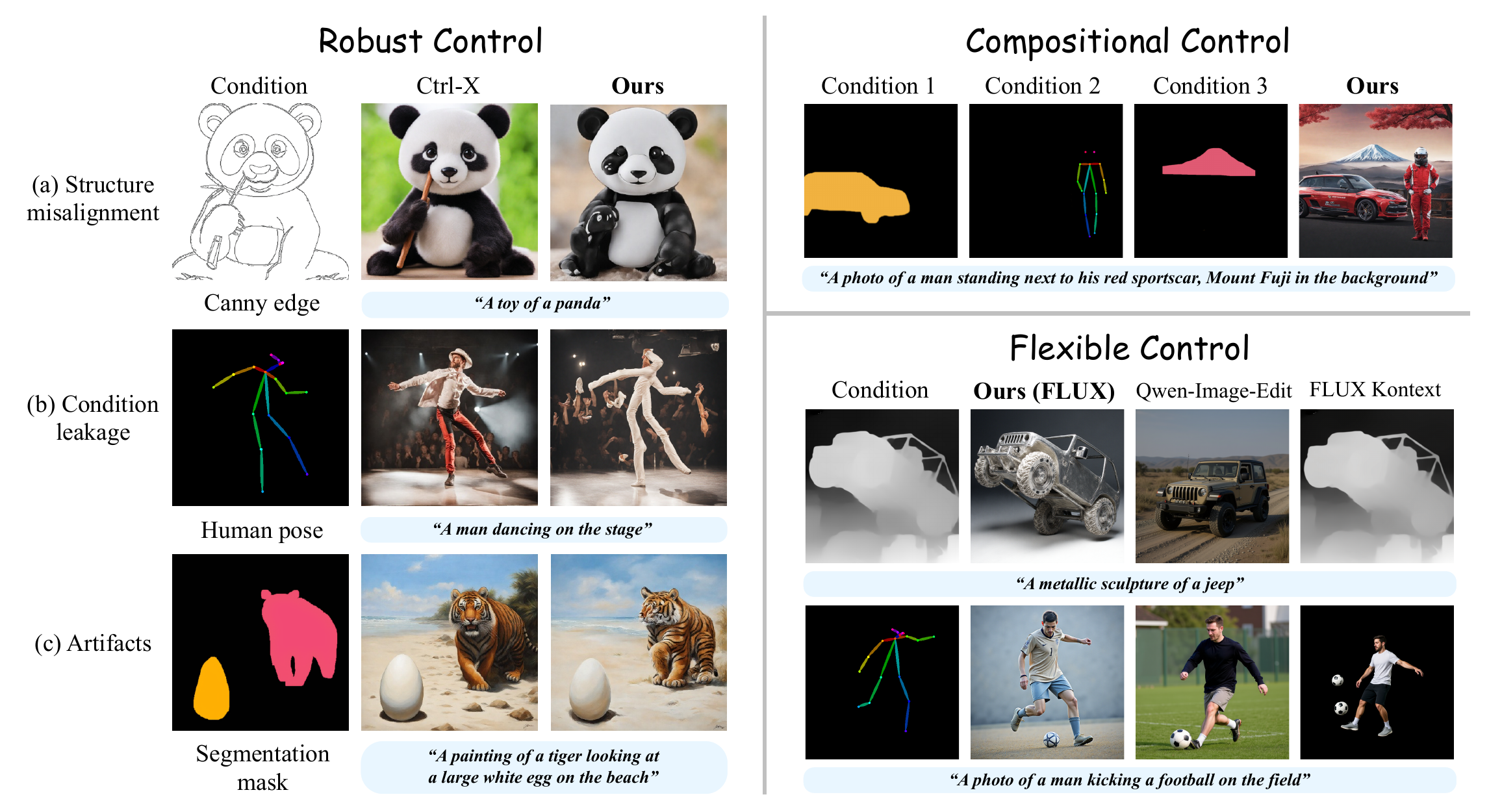}
        \captionof{figure}{\textbf{We propose a training-free framework that enables high-quality spatial control for pretrained text-to-image diffusion models under arbitrary spatial conditions.} (\textbf{\textit{Left}}) Our method addresses key limitations of prior training-free approaches to achieve more robust control; (\textbf{\textit{Top right}}) can handle compositional generation with multiple conditions and complex prompts; and (\textbf{\textit{Bottom right}}) can be flexibly extended to DiT-based architectures, achieving stronger structure control than training-based models like Qwen-Image-Edit~\citep{wu2025qwenimagetechnicalreport} and FLUX Kontext~\citep{labs2025flux1kontextflowmatching}.}
        \label{fig:teaser}
    \end{center} 
}]

\begin{abstract}
Text-to-image (T2I) diffusion models have shown remarkable success in generating high-quality images from text prompts. Recent efforts extend these models to incorporate conditional images (\eg, canny edge) for fine-grained spatial control. Among them, feature injection methods have emerged as a training-free alternative to traditional fine-tuning-based approaches. However, they often suffer from structural misalignment, condition leakage, and visual artifacts, especially when the condition image diverges significantly from natural RGB distributions. Through an analysis of existing methods, we identify a key limitation: the sampling schedule of condition features, previously unexplored,  fails to account for the evolving interplay between structure preservation and domain alignment throughout diffusion steps. Inspired by this observation, we propose a flexible training-free framework that decouples the sampling schedule of condition features from the denoising process, and systematically investigate the spectrum of feature injection schedules to achieve a better balance between structural alignment and appearance quality. We further enhance the sampling process by introducing a restart refinement schedule, and improve the visual quality with an appearance-rich prompting strategy. Together, these designs enable training-free controllable generation that is both structure-rich and appearance-rich. Extensive experiments demonstrate that our method achieves state-of-the-art performance under complex and diverse conditions. Owing to its generality, our framework naturally supports compositional conditional generation and generalizes across architectures in a plug-and-play manner, from UNet-based diffusion models to modern DiT backbones such as FLUX.
\end{abstract}

\section{Introduction} \label{sec:intro}

With the success of text-to-image (T2I) diffusion models, recent research has explored integrating conditional images, \eg, depth maps for spatial control. 
Early approaches, such as ControlNet~\citep{zhang2023controlNet}, rely on fine-tuning or auxiliary networks trained on paired data, which constrains their flexibility and scalability.
Recent studies have shown that the rich structural information encoded within diffusion features can be exploited to guide image generation without retraining, thereby enabling zero-shot control. They inject features extracted from the condition image at each timestep into the target image~\citep{hertz2022prompt-to-prompt, tumanyan2023plug-and-play, lin2024ctrlx, lin2025freecontrol}, and have shown promising performance across diverse conditioning scenarios.

However, these methods still possess several limitations, including structural misalignment, condition leakage, and visual artifacts (\cref{fig:teaser}). These issues become more pronounced when the condition image deviates significantly from natural RGB distributions, \eg, in pose or depth maps (\cref{fig:compare_sota_more}). This suggests that a key challenge lies in the domain gap between condition and natural image features in pretrained T2I diffusion models. We hypothesize that the injected condition features often lie outside the distribution of natural image features, which hinders the synthesis of high-fidelity results. This motivates us to analyze the temporal dynamics of diffusion features, observing a trade-off between structural fidelity and domain alignment (see \cref{fig:selfsim_vs_kldiv,fig:pca_results}).
These findings expose a fundamental limitation in existing training-free methods~\citep{hertz2022prompt-to-prompt, tumanyan2023plug-and-play, lin2024ctrlx}, which rely on condition features extracted at the \textit{same} timestep during denoising. This schedule fails to accommodate the evolving trade-off across timesteps: early features leads to loss of structural detail, while late features result in domain mismatch and condition leakage (\cref{fig:pca_results}). 

To address this, we generalize the sampling process of condition features and explore the design space of the feature injection schedule. The result shows that the optimal timestep is neither the same one as the target output image nor the latest one with the clearest features. Through a comprehensive investigation, we identify a family of candidate schedules that share an identical last timestep, among which a constant schedule yields consistently strong results. 
Building on these insights, we propose a more flexible feature injection framework that decouples the injection timestep from the denoising process. 
To further enhance control precision and visual fidelity, we apply a restart refinement schedule that iteratively mitigates visual artifacts introduced by injected features, and incorporate prompt augmentation to ensure semantic alignment with the condition image.
Together, these designs enable structure- and appearance-richer control of pretrained diffusion models. \cref{fig:pipeline} provides an overview of our framework, which consists of three key components: (i) \textit{Structure-Rich Injection (SRI)} injects condition features based on a principled sampling schedule; (ii) \textit{Restart Refinement (RR)} performs iterative forward–backward denoising; (iii) \textit{Appearance-Rich Prompting (ARP)} aligns the semantics of the appearance prompt with the condition image. 

Extensive experiments validate the effectiveness of our approach across diverse types of condition images, demonstrating improved structural consistency, visual fidelity, and semantic alignment compared to state-of-the-art training-free methods. 
In addition, our framework can readily handle compositional generation with multiple conditions and complex prompts. 
Thanks to its flexible and adaptable design, it seamlessly integrates with prior methods such as FreeControl~\citep{mo2023freecontrol} and extends to modern DiT backbones such as FLUX~\citep{flux2024}, surpassing even training-based controllable generation methods in structure fidelity, particularly under challenging conditions.

\section{Related Work}
\label{sec:related}

\subsection{T2I Diffusion Models}

Text-to-image (T2I) diffusion models~\citep{ho2022cascaded, saharia2022imagen, ramesh2022dalle2, rombach2022ldm, podell2023sdxl, esser2024sd3, flux2024} typically integrate textual information into the backbone~\citep{olaf2015unet, Peebles2023DiT} via cross-attention or classifier-free guidance~\citep{ho2022cfg}. In addition to architectural innovations, some work focuses on improving sampling quality and efficiency~\citep{song2020ddim, lu2022dpmsolver, karras2022edm, xu2023restart, liu2023rectifiedflow, song2023consistency, zhao2023unipc}. Our work explores different sampling strategies in the context of conditional text-to-image generation.

\subsection{Training-free Controllable Diffusion Models}
\label{subsec:tfcontrollable}

It is difficult to convey human preferences through text descriptions alone. Controllable diffusion models mitigate this problem by incorporating additional input signals to guide the generation process. Among these methods, training-based approaches train auxiliary modules or fine-tune the model. These methods are less relevant to our work and we leave the discussion to Supp. \cref{subsec:tbcontrollable}. In contrast, training-free controllable diffusion models operate at inference time to achieve condition control without additional training on task-specific paired data, allowing seamless generalization to new condition types and model checkpoints. These methods can be classified into three categories: \textbf{(i) Image editing}~\citep{cao2023masactrl, xu2024headrouter, epstein2023selfguidance, parmar2023pix2pix-zero, zhang2024samcontrol, tewel2025addit, jia2024designedit, couairon2022diffedit, Feng2025Dit4edit, dalva2024fluxspace, zhu2025kvedit, Avrahami2025StableFlow, wang2024rfedit, Xu2025unveil, xu2023infedit, wei2025freeflux, titov2024guideandrescale, hu2025islock} takes an input image and applies targeted modifications while preserving other regions of the image; \textbf{(ii) Image-to-image translation}~\citep{su2022ddib} learns mappings between images of different domains, with some works~\citep{alaluf2023crossimage, lin2024ctrlx, kwon2023diffuseit, huang2025attenst, go2024eyeforaneye, chung2024styleid} further conditioning the transformation on an additional appearance image; \textbf{(iii) Conditional text-to-image (T2I) generation} generates images consistent with both textual prompts and input control signals. These signals range from coarse layout constraints, such as bounding boxes~\citep{xiao2024rb, chen2023layoutguidance, xie2023boxdiff, li2025winwinlay, wang2025DADG}, semantic references like subject images~\citep{zhang2025freecus, feng2025personalizeanything, wang2025freeevent, ding2024freecustom, pham2024tale, rout2025rbmodulation}, to condition images that provide fine-grained control~\citep{lin2024ctrlx, mo2023freecontrol, tumanyan2023plug-and-play, lin2025freecontrol, bansal2023universalguidance, meng2021sdedit, kim2023densediffusion, lee2025scribbleguided}. As a \textbf{training-free conditional T2I method} conditioned on \textbf{condition images}, our approach extends this line of work by improving both control fidelity and visual quality across diverse conditions. To the best of our knowledge, we are also the first to demonstrate training-free structure control under multiple condition modalities on DiT-based architectures.

\section{Revisiting Sampling Schedules of Condition Features}
\label{sec:revisit}

\vspace{0.5em}
\noindent\textbf{Background.} Given a text prompt $\prompt$ and a condition image $\imagecond$ of arbitrary modality, the goal is to generate an output image $\imagegen$ that semantically aligns with the prompt $\prompt$ while preserving the structure of $\imagecond$. To this end, training-free approaches~\citep{tumanyan2023plug-and-play, mo2023freecontrol, lin2024ctrlx, lin2025freecontrol} typically leverage the diffusion features of a noisy latent $\latenttcond$ of the condition image.
Specifically, they encode $\imagecond$ into a clean latent $\latentzerocond$, and then obtain its noisy version $\latenttcond$ through DDIM inversion~\citep{song2020ddim} or the diffusion forward process. Intermediate features are subsequently extracted from the model backbone, and condition features are injected into those of $\imagegen$ at each timestep. We denote these condition features as $\featureltcond$, where $l$ refers to the layer index and $t$ denotes the timestep.

\vspace{0.5em}
\noindent\textbf{Limitations of Existing Methods.} While enabling zero-shot spatial control with diverse condition modalities, these methods often suffer from structural misalignment and condition leakage. For instance, Ctrl-X~\citep{lin2024ctrlx} fails to preserve the structure of the panda (\cref{fig:teaser}). Empirically, we observe that these failures are further exacerbated when the condition image deviates substantially from natural RGB images, as in the case of pose or depth maps shown in \cref{fig:compare_sota_more}. This suggests that a key challenge lies in the domain gap between the condition and natural image distributions in the feature space of pretrained diffusion models. We hypothesize that the injected features $\featureltcond$ fall outside the distribution of natural image features, thereby reducing their effectiveness in preserving the structure of the condition image $\imagecond$ during generation.

\begin{figure}[H]
    \centering
    \includegraphics[width=1.0\linewidth]{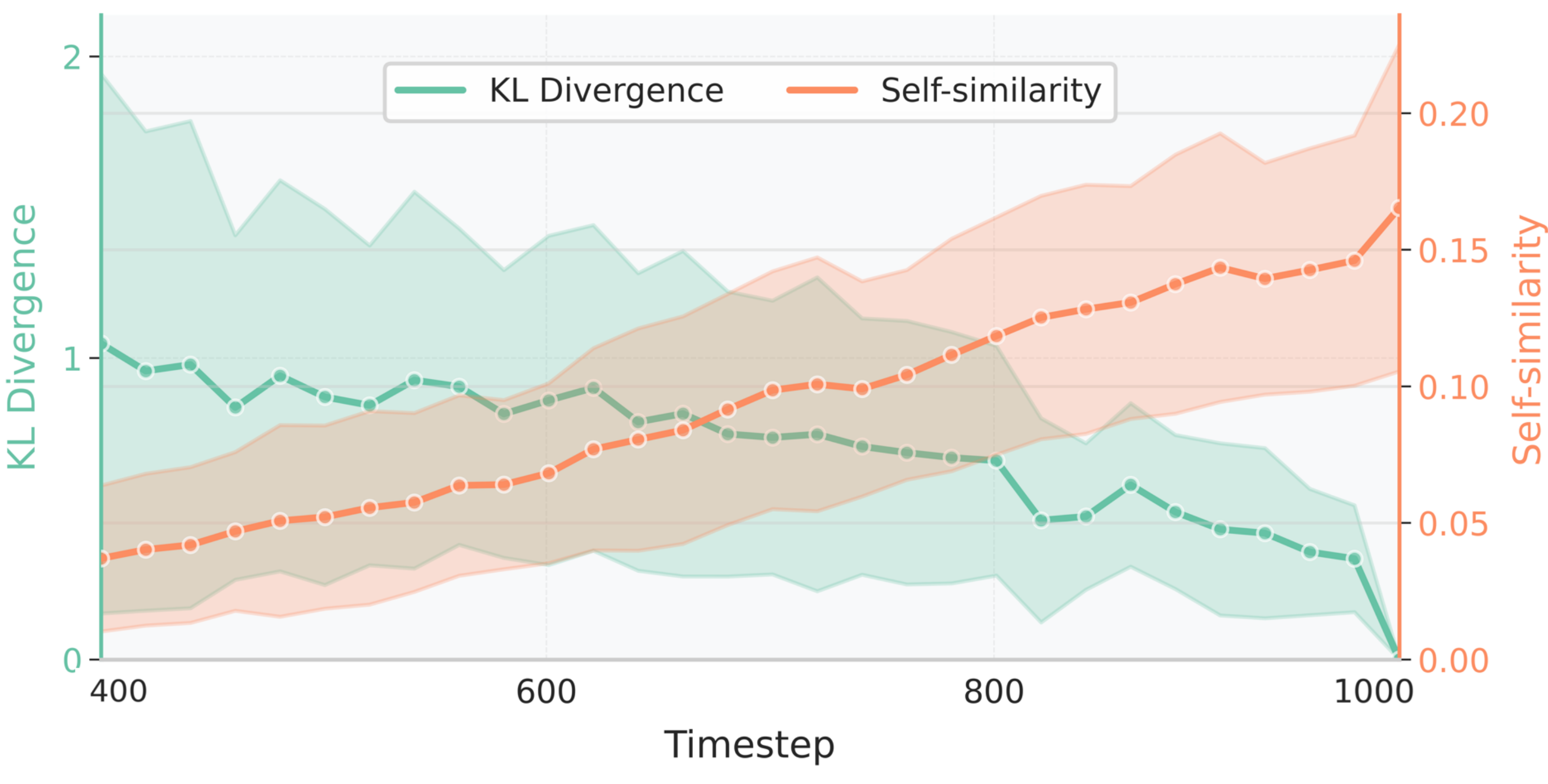}
    \caption{\textbf{The evolving curves of KL divergence and L2 distance of self-similarity matrices across diffusion timesteps.}}
    \label{fig:selfsim_vs_kldiv}
\end{figure}

\noindent\textbf{Analysis.} To validate this hypothesis, we quantitatively analyze features from 100 pairs of condition images across five common modalities (see Supp. \cref{supp:sec:analysis} for more details). As shown by the \textcolor{myorange}{orange} curve in \cref{fig:selfsim_vs_kldiv}, self-similarity distance decreases as noise is reduced, reflecting a progressive gain of fine-grained spatial cues. However, this improved structural fidelity comes at the cost of reduced domain alignment: the \textcolor{mygreen}{green} curve in \cref{fig:selfsim_vs_kldiv} shows that the KL divergence increases at lower timesteps, indicating a widening domain gap between natural and condition features. 

\begin{figure}[H]
    \centering
    \includegraphics[width=0.85\linewidth]{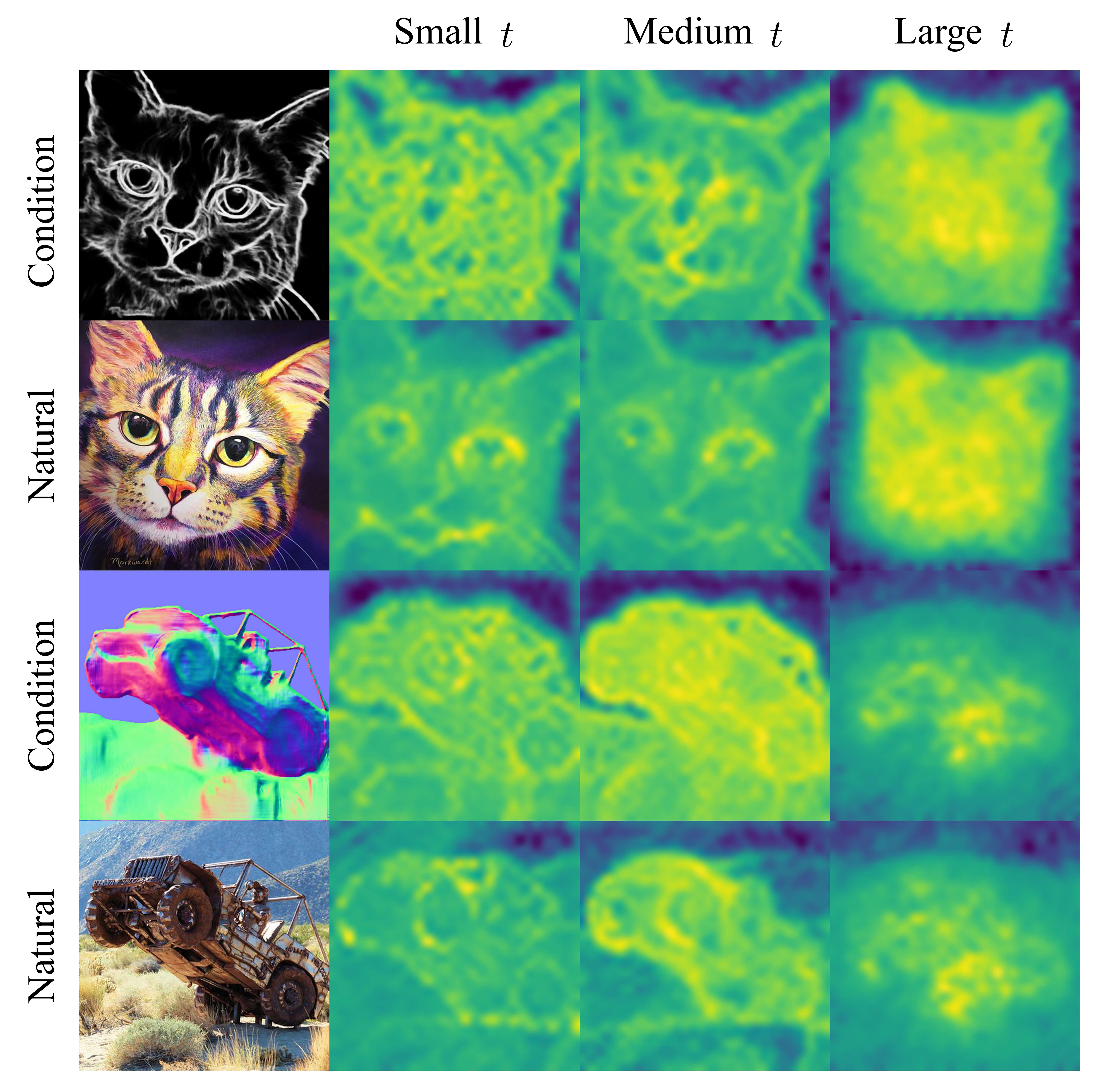}
    \caption{\textbf{Visualizing diffusion features extracted from the condition and natural images at various timesteps.}
    }
    \label{fig:pca_results}
\end{figure}

\begin{figure*}[t!]
  \centering
  \includegraphics[width=1.0\linewidth]{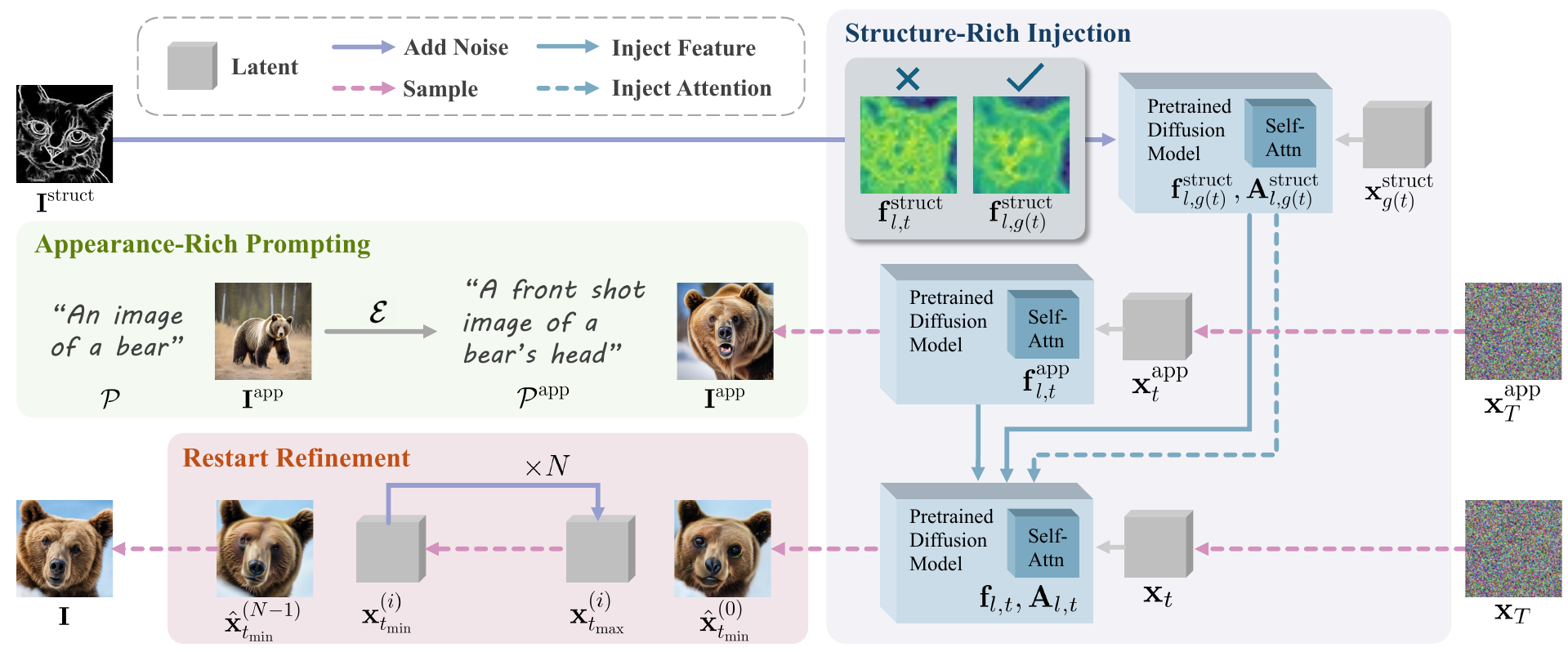}
  \caption{\textbf{Method overview.} 
  Given a condition image $\imagecond$ and a prompt $\prompt$, our method generates an output image $\imagegen$, aligning semantically with $\prompt$ while preserving the structure of $\imagecond$. Our framework consists of three key components. 
  (i) The \textbf{Structure-Rich Injection (SRI) module} (blue) injects structure-rich condition features $\featurelgtcond$ and attentions $\attnlgtcond$ into the output feature space to enable spatial control (Sec.~\ref{sec:method_feature}). 
  (ii) The \textbf{Restart Refinement (RR) module} (pink) iteratively adds noise to and denoises $\imagegen$ to refine visual details (Sec.~\ref{sec:method_restart}).
    (iii) The \textbf{Appearance-Rich Prompting (ARP) module}  (green) derives an enriched prompt $\promptapp$ based on the semantics of the condition image $\imagecond$ to generate a reference image $\imageapp$ for appearance transfer (Sec.~\ref{sec:method_prompt}).}
  \vspace{-1em}
  \label{fig:pipeline}
\end{figure*}

We also conduct principal component analysis and visualize the diffusion features in \cref{fig:pca_results}. There exists a visible discrepancy between the features of the condition image and its natural counterpart, which is more pronounced at smaller timesteps. Another notable pattern is that the primary structural information intended to be preserved emerges in the middle stage, while modality-specific details become prominent in the late stage. These observations highlight the limitations of previous methods: as the sampling schedule progresses, early features convey only coarse structural cues, whereas late features introduce out-of-distribution details in the output image, leading to structure misalignment and condition leakage.

\vspace{0.5em}
\noindent\textbf{Insight.} Motivated by these observations, we generalize the sampling schedule of condition features, decoupling it from the denoising process, and explore the function space of feature injection schedules. In this formulation, the structure latent from which the injected feature is extracted lies at timestep $g(t)$, with $g(t) = t$ covering the case in previous approaches as a special instance. We systematically evaluate the impact of the form of $g(t)$ on structure alignment and visual quality (see \cref{sec:SRI_ablation} for details) and summarize that (i) the optimal condition timestep is neither the output timestep nor the smallest one, but lies in the middle stage; and
(ii) schedules with their last timestep anchored around medium timesteps consistently deliver the optimal balance between structural fidelity and visual quality, largely independent of their functional form.
\vspace{-0.5em}
\section{Method} \label{sec:method}

Building on the insights in \cref{sec:revisit}, we introduce a training-free controllable T2I generation framework that enables flexible, structure- and appearance-richer control.
Our approach comprises three components, as illustrated in \cref{fig:pipeline}: (i) \textbf{Structure-Rich Injection (SRI)} injects condition features based on a principled sampling schedule, with a better balance of structure preservation and domain alignment (\cref{sec:method_feature}); 
(ii) \textbf{Restart Refinement (RR)} schedule performs iterative refinement to suppress visual artifacts and improve output fidelity (\cref{sec:method_restart}).
(iii) \textbf{Appearance-Rich Prompting (ARP)} enriches the original prompt $\prompt$ informed by $\imagecond$, facilitating appearance guidance (\cref{sec:method_prompt}). 
Together, these modules enable structure- and appearance-aware generation across diverse conditions, all in a zero-shot manner. We now describe each component in detail.

\subsection{Structure-Rich Injection} \label{sec:method_feature}

The structure-rich injection strategy adopts a sampling schedule where the extracted condition features are both semantically compatible and structurally informative. Specifically, we begin by encoding the structure condition image $\imagecond$ using the pretrained model to obtain the condition features and attention maps, as illustrated in \cref{fig:pipeline}. Prior work \citep{tumanyan2023plug-and-play, mo2023freecontrol, lin2024ctrlx, lin2025freecontrol} typically extracts condition features $\featureltcond$ and attention maps $\attnltcond$ at a self-attention layer $l$ and timestep $t$, and injects them into the generation branch by substituting the corresponding $\featureltgen$ and $\attnltgen$ at the same layer and timestep (the bottom branch in \cref{fig:pipeline}). We use "injection" to refer to this substitution of features and/or attention maps throughout the paper.

According to the feature analysis in \cref{sec:revisit}, however, we select features from a separate schedule $g(t)$, where $g(\cdot)$ is a general function of the current timestep $t$. As shown in the blue block of \cref{fig:pipeline}, the extracted features $\featurelgtcond$ and attention maps $\attnlgtcond$ are then used to replace their counterparts $\featureltgen$ and $\attnltgen$ in the generation backbone at timestep~$t$: $\featureltgen \gets \featurelgtcond \mathrm{~~and~~} \attnltgen \gets \attnlgtcond.$

\vspace{0.5em}
\noindent\textbf{Single-Step Sampling and Caching.} 
Guided by the insights in \cref{sec:revisit}, we adopt a simple yet effective schedule $g(t)=600$ for subsequent experiments. Since the features and attention maps of the condition image need to be computed only once, they can then be cached and reused throughout the denoising process, resulting in an improved computational efficiency of fine-grained structure control. Please refer to \cref{sec:exp} and Supp. \cref{supp:subsec:inference} for detailed results.

\subsection{Restart Refinement} \label{sec:method_restart}
Injection-based approaches can inherently introduce out-of-distribution artifacts and condition leakage during denoising. To address this, we adopt a restart refinement schedule inspired by diffusion-based sampling methods~\citep{xu2023restart}.
As illustrated in the pink block of \cref{fig:pipeline}, after structure and appearance control, we inject noise at an intermediate timestep $\restarttmin$, effectively restarting the denoising process by transitioning the latent to $\restarttmax$ step. A DDIM backward step is then applied. This forward–backward cycle is repeated $\restarttimes$ times within $[\restarttmin, \restarttmax]$. In the $i^{\textrm{th}}$ iteration ($i \in \{0, \dots, \restarttimes-1\}$), the restart proceeds as follows:
\begin{align}
\small
(\textrm{Forward}) \, \latenttmaxiplusoneth &= \latenttminith + \noisetmintotmax,\quad
\label{eqn:restart_ddim} \\
\small
(\textrm{Backward}) \, \latenttminiplusoneth &= h_{\mathrm{DDIM}}(\latenttmaxiplusoneth, \nonumber \restarttmax \to \restarttmin), 
\end{align}
where the initial $\latenttzeroith$ is obtained by simulating the DDIM step until $\restarttmin$: $\latenttzeroith = h_{\mathrm{DDIM}}(\latentTgen, T \to \restarttmin)$, and the noise $\noisetmintotmax$ is sampled from the Gaussian forward perturbation kernel induced by the base model scheduler from $\restarttmin$ to $\restarttmax$. Through this schedule, our approach achieves better visual fidelity as demonstrated in \cref{fig:RR_ablation}.

\subsection{Appearance-Rich Prompting}  \label{sec:method_prompt}
To enhance the realism of the generated image, prior work~\citep{mo2023freecontrol, lin2024ctrlx} generates an appearance image $\imageapp$ and performs appearance transfer (the middle branch in \cref{fig:pipeline}). However, the semantic misalignment between the condition image and coarse user prompts can hinder the establishment of semantic correspondence between the appearance image $\imageapp$ and the output image $\imagegen$ in existing methods, leading to artifacts. For example, as illustrated in \cref{fig:pipeline}, the condition image is a frontal view of a cat's head, while the text prompt specifies "a bear", resulting in a semantically misaligned appearance image $\imageapp$ that depicts a full-body bear. 

To tackle this issue, we propose Appearance-Rich Prompting (ARP), a strategy that leverages multimodal large language models~\citep{achiam2023gpt4} to systematically align the semantics of $\prompt$ with those of the conditions $\imagecond$, as shown in the green block of \cref{fig:pipeline}. Examples are provided in the ablation study in \cref{fig:ARP_ablation} and details of the prompt transform $\mathcal{E}$ are provided in Supp. \cref{supp:sec:method}.
\section{Experiments} \label{sec:exp}

\subsection{Setup}

\noindent\textbf{Dataset.} We construct our evaluation dataset based on datasets from prior work~\citep{mo2023freecontrol, lin2024ctrlx}. Specifically, we use 7 common condition types: canny edges, depth maps, normal maps, human poses, segmentation masks, HED edges, and scribble drawings, and form a balanced dataset with over 20 image-text pairs for each type (see Supp. \cref{supp:sec:details} for details).  

\noindent\textbf{Baselines.} We evaluate our method against 6 existing training-free baselines: Ctrl-X~\citep{lin2024ctrlx}, FreeControl~\citep{mo2023freecontrol}, PnP~\citep{tumanyan2023plug-and-play}, P2P~\citep{hertz2022prompt-to-prompt}, SDEdit~\citep{meng2021sdedit}, and InfEdit~\citep{xu2023infedit}, as well as 2 training-based baselines: ControlNet~\citep{zhang2023controlNet} and T2I-Adapter~\citep{mou2023t2i}.
Experiment results on the condition types supported by T2I-Adapter are provided in Supp. \cref{supp:subsec:quantitative}. Wherever possible, we implement each method using SDXL 1.0~\citep{podell2023sdxl}; otherwise, we use their best-performing publicly available checkpoints. We set inversion and denoising steps to 50 for all baselines.

\noindent\textbf{Evaluation Metrics.} 
Following prior work~\citep{mo2023freecontrol,lin2024ctrlx}, we employ three widely-adopted metrics for comparison: CLIP~\citep{radford2021clip} for text-image alignment, DINO self-similarity (Self-sim)~\citep{caron2021dino, tumanyan2022splicing} for structural alignment, and Condition LPIPS~\citep{zhang2018lpips} for appearance quality. See Supp. \cref{supp:sec:details} for details of these evaluation metrics. To ensure a comprehensive evaluation, we also incorporate three reward models that better capture human preferences~\citep{fu2023dreamsim, xu2023imagereward, wu2023hpsv2, ma2025hpsv3}: DreamSim~\citep{fu2023dreamsim} for perceptual similarity, ImageReward~\citep{xu2023imagereward} for text alignment and image quality, and HPSv2~\citep{wu2023hpsv2} for human-aligned overall generation quality. All experiments were repeated three times and we report the average results.

\begin{figure}[t]
\centering{
  \includegraphics[width=\linewidth]{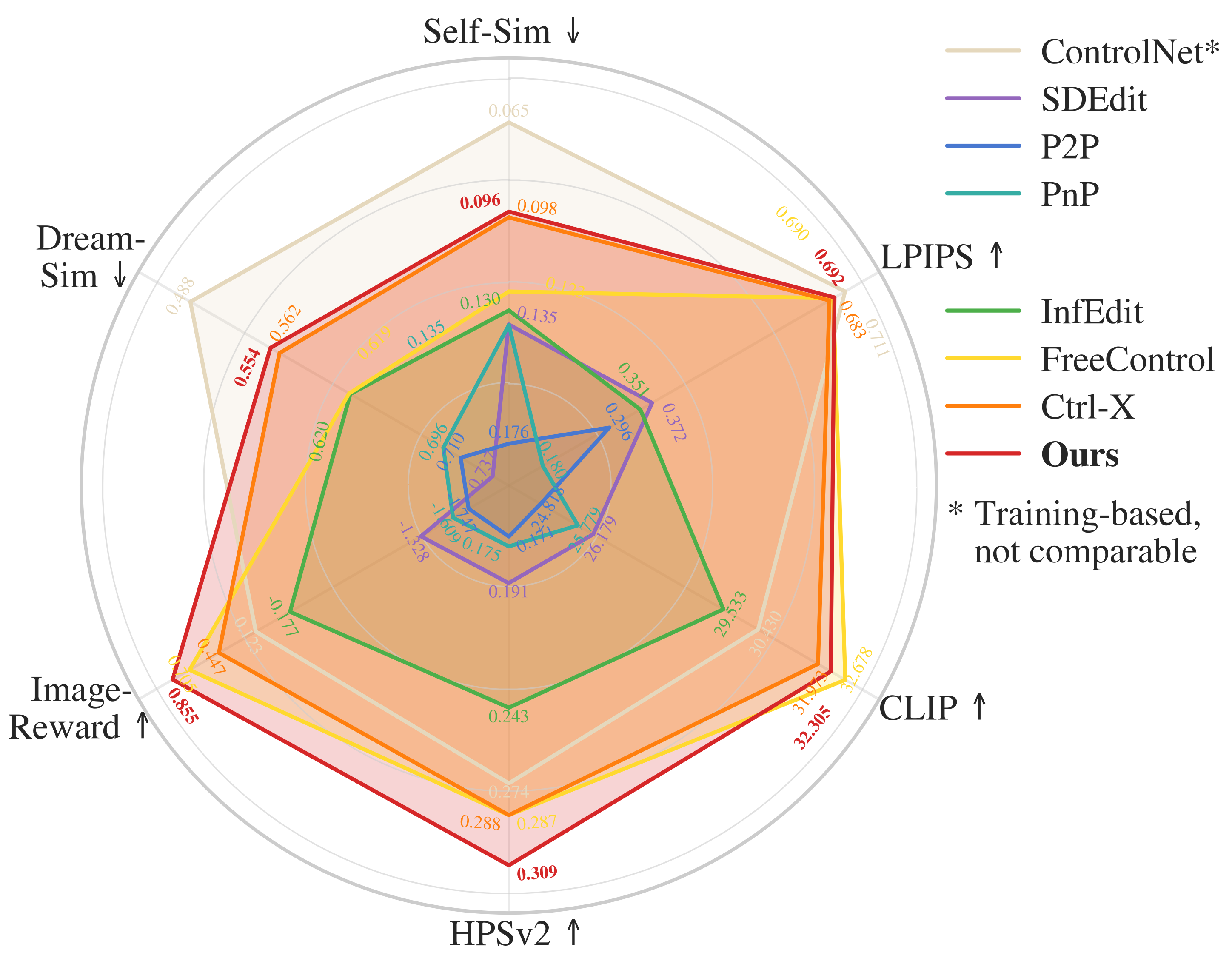}  \caption{\textbf{Our method achieves SOTA performance among all training-free methods}; in the radar chart, greater distance from the center indicates superior results.}
  \label{fig:radar_chart}
  }
\end{figure}

\begin{figure*}[t]
  \centering
  \includegraphics[width=0.95\linewidth]{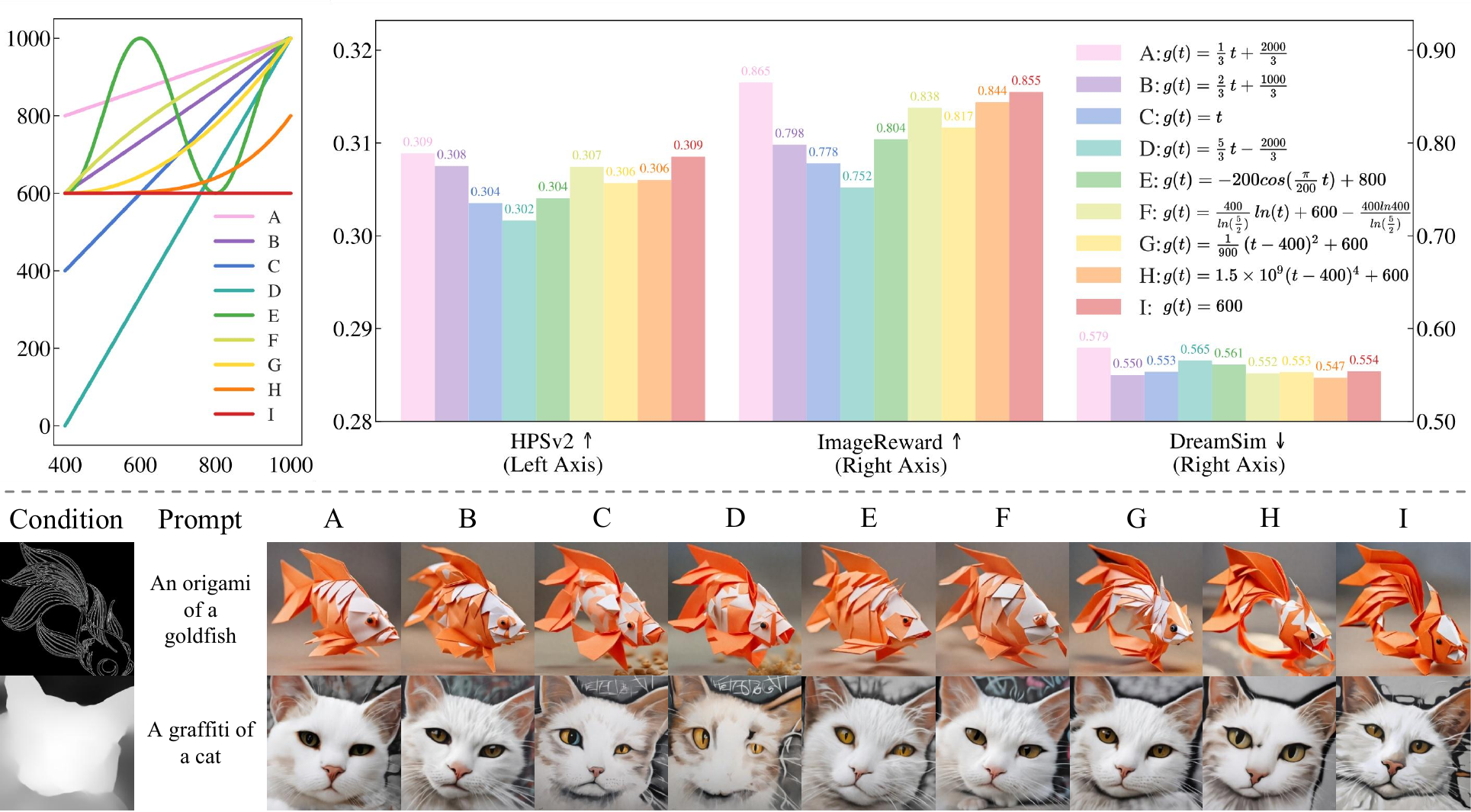}
  \caption{\textbf{Ablation of SRI schedules}. We report quantitative (\textbf{\textit{Top}}) and qualitative (\textbf{\textit{Bottom}}) results for different injection schedules.}
  \label{fig:ablation_injection}
\end{figure*}

\subsection{Comparison with \ac{sota}}

\noindent\textbf{Analysis.} \cref{fig:compare_sota_more,fig:radar_chart} present quantitative and qualitative comparisons between our method and the baselines, respectively. While training-based approaches like ControlNet~\citep{zhang2023controlNet} and T2I-Adapter~\citep{mou2023t2i} exhibit lower Self-sim scores, they often fail to adhere to the text prompts (\eg, \textit{origami} in \cref{fig:compare_sota_more}), leading to impaired text-image alignment. In contrast, our method achieves robust structural alignment while maintaining superior text-image consistency.

Training-free baselines also exhibit several limitations. SDEdit~\citep{meng2021sdedit}, PnP~\citep{tumanyan2023plug-and-play}, and P2P~\citep{hertz2022prompt-to-prompt} are prone to condition leakage, producing outputs that closely resemble the condition image. FreeControl~\citep{mo2023freecontrol} and InfEdit~\citep{xu2023infedit} often yield results with inferior structural alignment and artifacts (high DreamSim and DINO self-similarity). While Ctrl-X~\citep{lin2024ctrlx} performs reasonably well in many cases, it still suffers from structural misalignment (column 5), condition leakage (column 3), and artifacts (columns 1, 2, and 4), as shown in \cref{fig:compare_sota_more}. In contrast, our method consistently outperforms these baselines in structural preservation, text-image alignment, and visual quality, excelling in difficult scenarios such as abstract conditions (\eg, pose), multi-object scenes, and challenging prompts (\eg, \textit{origami}). Quantitative evaluations further confirm the advantage of our approach. As shown in \cref{fig:radar_chart}, our method surpasses training-free baselines across nearly all metrics. Please refer to \cref{fig:quality_result} and \cref{tab:sota} in the supplementary materials for additional qualitative and quantitative results.

\begin{table}[h]
  \centering
  \small
  \setlength{\tabcolsep}{1mm}
  \renewcommand{\arraystretch}{1.15}
  \begin{threeparttable}
    \caption{\textbf{Computational efficiency and user study.}}
    \vspace{-1em}
    \label{tab:compute_and_user}
    \begin{tabular*}{\linewidth}{@{\extracolsep{\fill}}@{} l c c  | c @{}}
      \toprule
      & \makecell[c]{Time (s)}
      & \makecell[c]{Memory (GB)}
      & \makecell[c]{Preference Rate} \\
      \midrule
      InfEdit~\citep{xu2023infedit}        & 31.84  & 52.44 & 11.42\% \\
      FreeControl~\citep{mo2023freecontrol} & 781.57 & 55.46 & 10.67\% \\
      Ctrl-X~\citep{lin2024ctrlx}          & 19.37  & 18.79 & 21.67\% \\
      \textbf{Ours}                        & \textbf{18.79} & \textbf{18.77} & \textbf{56.25\%} \\
      \bottomrule
    \end{tabular*}
    \vspace{-1em}
  \end{threeparttable}
\end{table}

\begin{figure*}[t]
  \centering
  \includegraphics[width=0.92\linewidth]{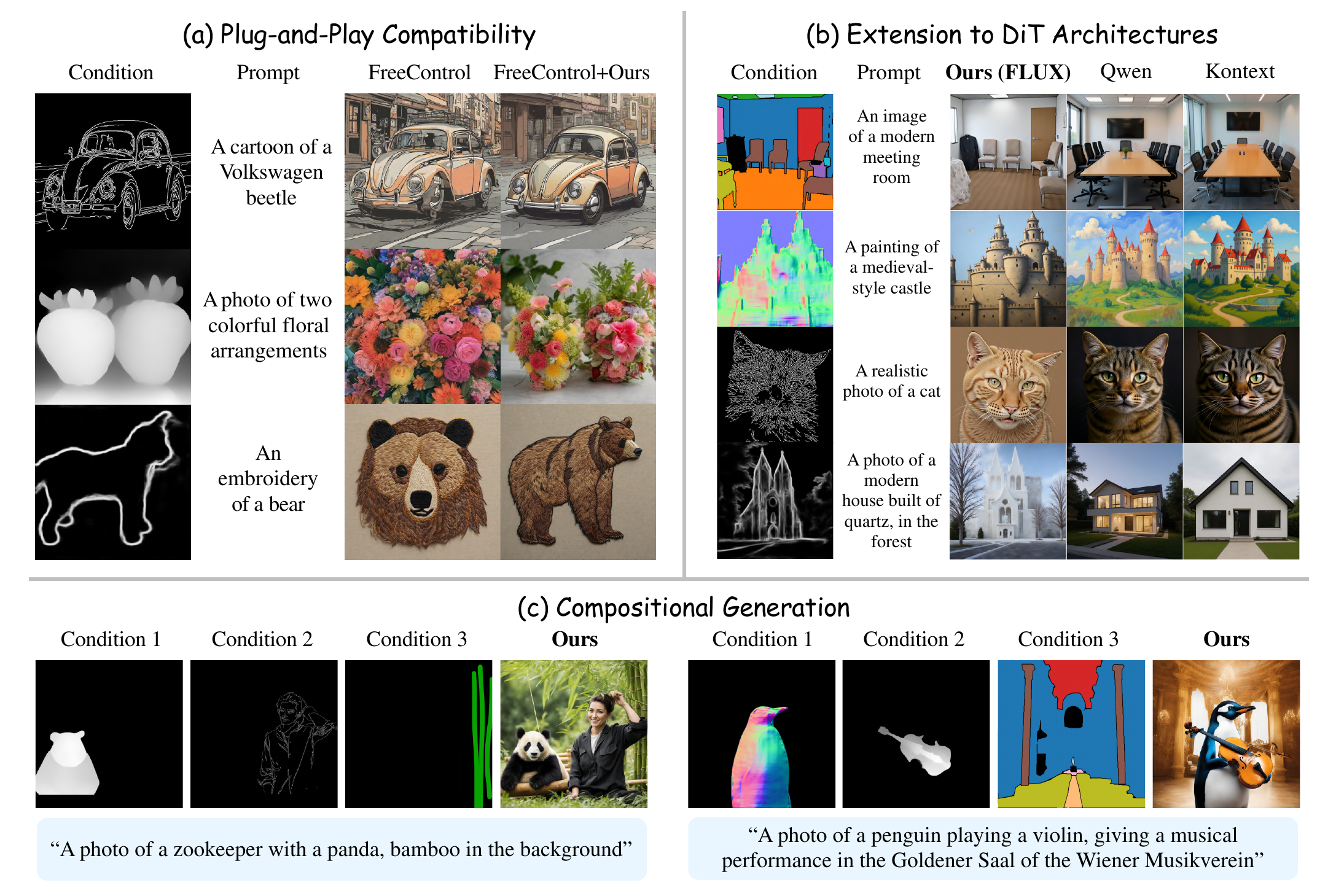}
    \caption{\textbf{Flexibility of our framework.} (a) As a plug-in, our method enhances FreeControl~\citep{mo2023freecontrol} with stronger structure preservation and visual quality. (b) Our method can be flexibly extended to the DiT-based FLUX backbone~\citep{flux2024}. (c) Our method can handle compositional conditions and complex prompts in a single generation.}
  \label{fig:flexible_exp}
\end{figure*}

\noindent\textbf{User Study. }We further conduct human evaluations to validate the effectiveness of our framework. We compare our method against three strongest baselines: InfEdit~\citep{xu2023infedit}, FreeControl~\citep{mo2023freecontrol} and Ctrl-X~\citep{lin2024ctrlx}. \cref{tab:compute_and_user} shows the result of the human evaluation, where 56.25\% of participants prefer the results produced by our method, highlighting the effectiveness of our approach. Please refer to Supp. \cref{supp:sec:details} for details on the user study execution.

\vspace{0.5em}
\noindent\textbf{Computational Efficiency. }
While the RR module introduces slightly more sampling steps, our single-step sampling and caching strategy effectively eliminates redundant feature computations, ensuring high efficiency. \cref{tab:compute_and_user} reports the average inference time and memory used by the 4 strongest methods on a single A800 80G GPU. Our method achieves the fastest inference speed and a relatively low memory cost, confirming the computational efficiency of our method. The time consumption of each module within our method is provided in Supp. \cref{supp:sec:exp}.

\subsection{Ablation Study}

\noindent\textbf{Structure-Rich Injection.} \label{sec:SRI_ablation} 
We explore different forms of injection schedules $g(t)$ and evaluate their impact on structure alignment and visual fidelity. The results are shown in \cref{fig:ablation_injection}. 
Our ablation on linear schedules (\cref{fig:ablation_injection}, A–D) reveals a clear trade-off: schedules biased toward larger timesteps (A) weaken structural alignment and increase DreamSim scores, whereas those biased toward smaller timesteps (C–D) degrade both visual quality (lower HPSv2 and ImageReward scores) and structural fidelity due to excessive modality-specific cues. In contrast, the medium-timestep schedules (B) strike a favorable balance, yielding stronger structure preservation without sacrificing visual realism.

Beyond linear functions, we further examine a family of schedules (E–I) that share the same final timestep but differ in monotonicity, convexity, and initialization. All of them deliver consistently strong performance, suggesting robustness to functional variations. Specifically, a constant-timestep schedule consistently achieves competitive results, providing a simple yet effective alternative for fine-grained structure control and domain alignment. We detail the protocol for selecting these functions in Supp. \cref{supp:sec:details}.

\begin{table}[t]
\centering
\small
\caption{\textbf{Quantitative ablation of ARP and RR.}}
\vspace{-1em}
\label{tab:ARP_RR_ablation}
\begin{tabular}{l c c c}
\toprule
& Dream-Sim $\downarrow$ & Image-Reward $\uparrow$ & HPSv2 $\uparrow$ \\
\midrule
w/o ARP       & 0.558 & 0.799 & 0.308 \\
w/o RR        & \textbf{0.544} & 0.518 & 0.286 \\
\textbf{Ours} & 0.554 & \textbf{0.855} & \textbf{0.309} \\
\bottomrule
\end{tabular}
\end{table}

\begin{figure}[htbp]
  \centering
  \includegraphics[width=0.8\linewidth]{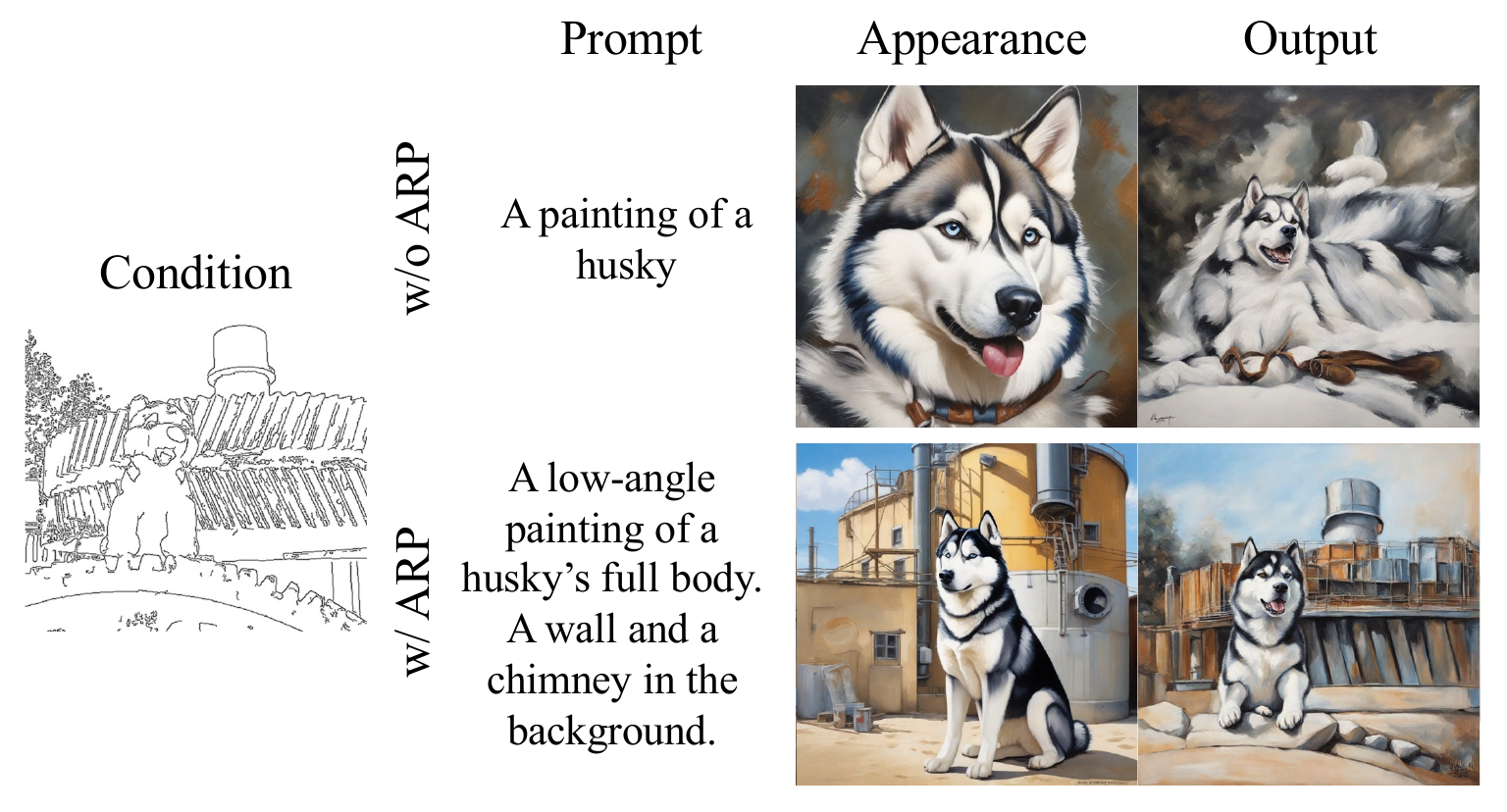}
  \caption{\textbf{Ablation of Appearance-Rich Prompting (ARP)}. See \cref{fig:supp_ablation_prompt} in the supplementary materials for more cases.}
  \label{fig:ARP_ablation}
\end{figure}

\vspace{0.5em}
\noindent\textbf{Appearance-Rich Prompting.} \cref{fig:ARP_ablation} demonstrates that the ARP module effectively adapts the prompt to capture key visual attributes of the condition image. \cref{tab:ARP_RR_ablation} further shows that removal of the ARP module decreases performance in all three metrics, verifying its effectiveness in enhancing structural preservation and visual fidelity.

\begin{figure}[htbp]
  \centering
  \includegraphics[width=0.8\linewidth]{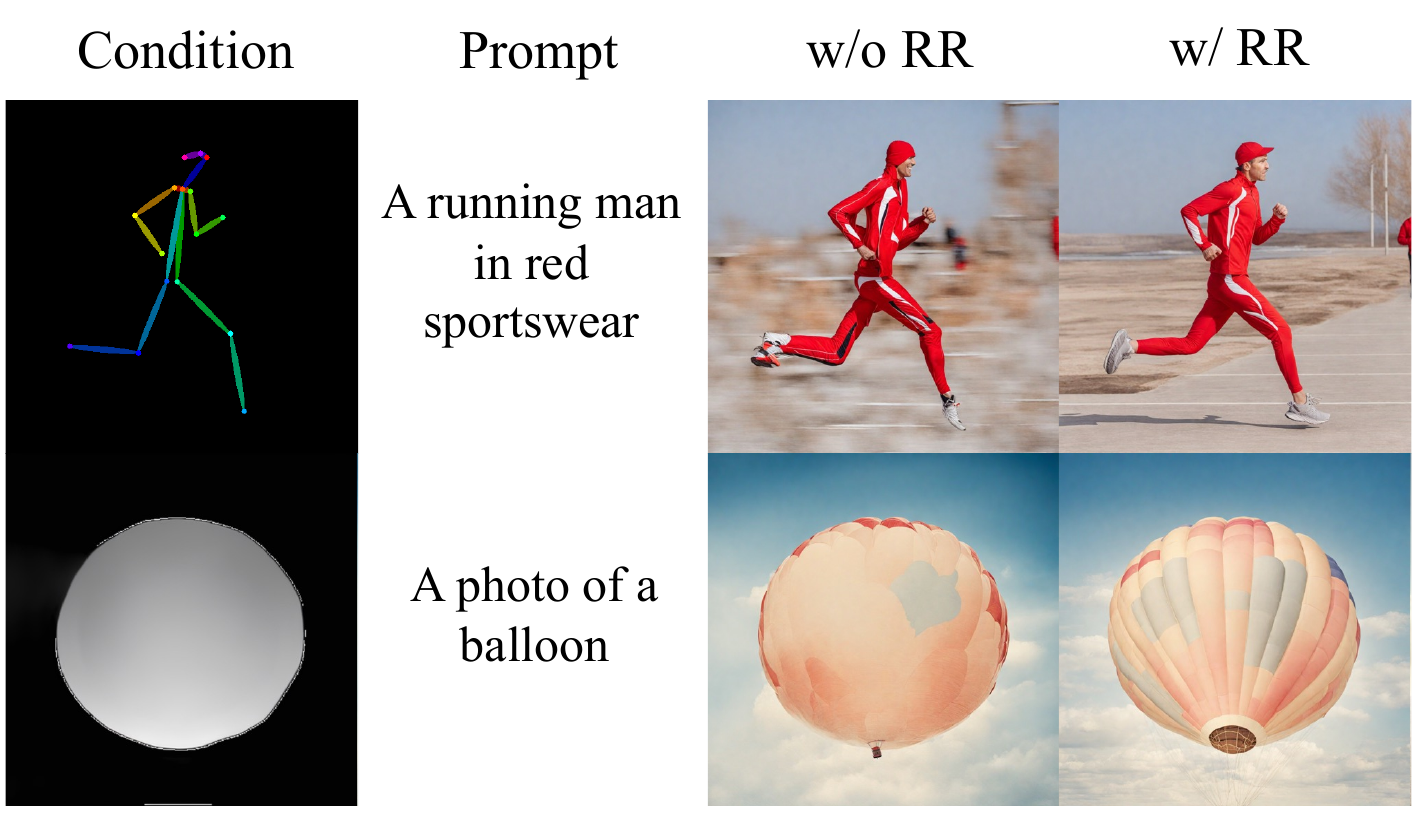}
  \caption{\textbf{Ablation of Restart Refinement (RR)}. See \cref{fig:supp_ablation_restart} in the supplementary materials for more cases.}
  \label{fig:RR_ablation}
\end{figure}

\vspace{0.5em}
\noindent\textbf{Restart Refinement.} As shown in \cref{fig:RR_ablation}, the RR schedule mitigates both condition leakage and artifacts, leading to improved generation quality while maintaining strong structural alignment. 

\subsection{Versatility and Extensibility}

\noindent\textbf{Plug-and-Play Compatibility.} Our method can be seamlessly integrated into existing conditional T2I approaches to boost their generation performance. To demonstrate this, we applied our framework as a plug-in module for FreeControl~\citep{mo2023freecontrol}. As shown in \cref{fig:flexible_exp}~(a), this integration significantly improves both structural preservation and perceptual quality, confirming the versatility and practical value of our proposed approach.

\vspace{0.5em}
\noindent\textbf{Extension to DiT Architectures.}
Since our framework relies on model-agnostic components for injecting structural context, it can be readily extended beyond traditional UNet backbones to Diffusion Transformer (DiT)~\cite{Peebles2023DiT} architectures. We integrate our method with the popular DiT-based framework FLUX~\citep{flux2024} and compare our results against two \textbf{training-based} large image generation models built on DiT architectures: Qwen-Image-Edit~\citep{wu2025qwenimagetechnicalreport} and FLUX Kontext~\citep{labs2025flux1kontextflowmatching}. As shown in \cref{fig:flexible_exp}~(b), although our method is \textbf{training-free}, it achieves superior structural control and generation quality across diverse conditions, whereas the two large models often fall short in structural alignment. Please refer to Supp. \cref{supp:sec:details} for more implementation details.
\section{Conclusion} \label{sec:conclusion}
We propose a training-free framework for conditional text-to-image generation. By leveraging the features of pretrained diffusion models in a principled manner, our method balances structural fidelity and appearance quality while automatically enhancing image realism and prompt relevance. Our investigation facilitates the understanding of the feature space of T2I diffusion models and achieves a strong, general, and robust solution for injection-based pipelines. 

\vspace{0.5em}
\noindent\textbf{Limitations and future directions.} Although we draw meaningful conclusions from principled investigations and design an effective method, it remains a promising future direction to interpret the results theoretically. A formal explanation in a high-dimensional feature space is a non-trivial task, and it requires further dedication from the research community.

{
    \small
    \bibliographystyle{ieeenat_fullname}
    \bibliography{reference_header,main_arxiv}
}

\clearpage
\appendix

We provide preliminaries in \cref{supp:sec:pre} and additional related work discussing training-based controllable diffusion methods in \cref{subsec:tbcontrollable}. In \cref{supp:sec:analysis}, we further analyze the domain gap and structure preservation of diffusion features. Then we elaborate on the implementation details of our proposed method in \cref{supp:sec:method} and the experimental setups in \cref{supp:sec:details}. We show additional experimental results in \cref{supp:sec:exp}. We discuss broader societal impacts of the work in \cref{supp:sec:broader}.

\section{Preliminaries}
\label{supp:sec:pre}

\noindent\textbf{Diffusion Models. }Diffusion models are a family of probabilistic generative models characterized by two processes.

The \emph{forward process} iteratively adds Gaussian noise to a clean image $\latentzero$ to obtain $\latentt$ for time step $t \sim [1, T]$, which can be reparameterized in terms of a noise schedule $\alpha_t$ where
\begin{equation} \label{eqn:diff_forward}
    \mathbf{x}_t = \sqrt{\alpha_t} \mathbf{x}_0 + \sqrt{1 - \alpha_t} \mathbf{\epsilon},
\end{equation}
for $\mathbf{\epsilon} \sim \mathcal{N}(0, \mathbb{I})$.

The \emph{backward process} generates images by iteratively denoising an initial Gaussian noise $\mathbf{x}_T \sim \mathcal{N}(0, \mathbb{I})$, also known as diffusion sampling \citep{ho2020ddpm}. This process uses a parameterized denoising network $\mathbf{\epsilon}_\theta$ conditioned on a text prompt $\prompt$, where at time step $t$ we obtain a cleaner $\mathbf{x}_{t - 1}$ as
\begin{equation} \label{eqn:diff_backward}
    \mathbf{x}_{t - 1} = \sqrt{\alpha_{t - 1}} \hat{\mathbf{x}}_t + \sqrt{1 - \alpha_{t - 1}} \mathbf{\epsilon}_\theta(\mathbf{x}_t \mid t, \prompt) ,
\end{equation}
\begin{equation}
    \hat{\mathbf{x}}_t = \frac{\mathbf{x}_t - \sqrt{1 - \alpha_t} \mathbf{\epsilon}_\theta(\mathbf{x}_t \mid t, \prompt)}{\sqrt{\alpha_t}} .
\end{equation}
Intuitively, $\hat{\mathbf{x}}_{t}$ approximates the initial clean image, which is subsequently perturbed with an appropriate amount of noise to produce the input for the following timestep.

\vspace{0.5em}
\noindent\textbf{Guidance.} The iterative inference of diffusion enables people to guide the sampling process on auxiliary information. \emph{Guidance} modifies \cref{eqn:diff_backward} to compose additional score functions that point toward richer and specifically conditioned distributions \citep{bansal2023universalguidance, epstein2023selfguidance}, expressed as
\begin{equation} \label{eqn:guidance}
    \hat{\mathbf{\epsilon}}_\theta(\mathbf{x}_t \mid t, \prompt) = \mathbf{\epsilon}(\mathbf{x}_t \mid t, \prompt) - s \, \mathbf{g}(\mathbf{x}_t \mid t, y) ,
\end{equation}
where $\mathbf{g}$ is an energy function and $s$ is the guidance strength. In practice, $\mathbf{g}$ can range from classifier-free guidance (where $\mathbf{g} = \mathbf{\epsilon}$ and $y = \emptyset$, \ie the empty prompt) to improve image quality and prompt adherence for T2I diffusion \citep{ho2022cfg, rombach2022ldm}, to arbitrary gradients computed from auxiliary models or diffusion features common to guidance-based controllable generation \citep{bansal2023universalguidance, epstein2023selfguidance, mo2023freecontrol}. Thus, guidance provides the customizability on the type and variety of conditioning for controllable generation, as it merely requires a differentiable loss with respect to $\mathbf{x}_t$. However, the need for backpropagation during inference often leads to increased memory consumption and slower inference speed. Moreover, guidance-based methods often fail to capture fine structural details in controllable generation tasks. 

\vspace{0.5em}
\noindent\textbf{Diffusion model features.} 
Let $\mathbf{f}_{l,t} \in \mathbb{R}^{HW \times c}$ be the diffusion feature with height $H$, width $W$, and channel size $c$ at time step $t$ right before attention layer $l$. Then, the self-attention operation is
\begin{equation}
\begin{gathered}
    \mathbf{Q} = \mathbf{f}_{l, t} \mathbf{W}^Q_l, \quad
    \mathbf{K} = \mathbf{f}_{l, t} \mathbf{W}^K_l, \quad
    \mathbf{V} = \mathbf{f}_{l, t} \mathbf{W}^V_l , \\
    \mathbf{f}_{l, t} \leftarrow \mathbf{A} \mathbf{V}, \quad
    \mathbf{A} = \mathrm{softmax}\left( \frac{\mathbf{Q} \mathbf{K}^\top}{\sqrt{d}} \right) ,
\end{gathered}
\end{equation}
where $\mathbf{W}^Q_l, \mathbf{W}^K_l, \mathbf{W}^V_l \in \mathbb{R}^{c \times d}$ are linear transformations which produce the query $\mathbf{Q}$, key $\mathbf{K}$, and value $\mathbf{V}$, respectively, and $d$ is the dimensionality of the attention space. The $\mathrm{softmax}$ operation is applied across the second $(HW)$-dimension (typically, $c = d$ in diffusion models). Intuitively, the attention map $\mathbf{A} \in \mathbb{R}^{(HW) \times (HW)}$ encodes how each pixel in $\mathbf{Q}$ corresponds to each in $\mathbf{K}$, which then rearranges and weighs $\mathbf{V}$. The rich structural information embedded in features of pretrained diffusion models lays the foundation for extensive training-free controllable generation approaches, and, together with the common issues of training-free methods, motivates us to study the temporal dynamics of diffusion features.

\begin{figure*}[t]
  \centering
  \includegraphics[width=1.0\linewidth]{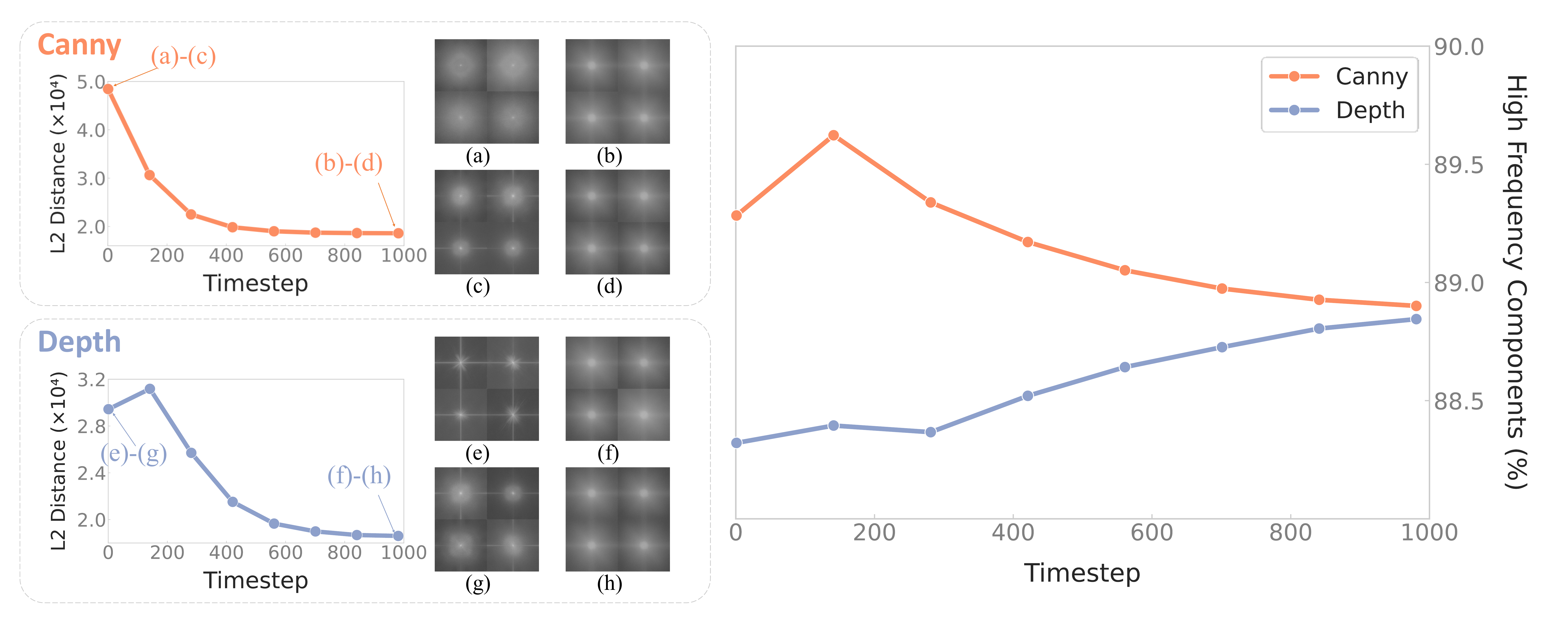}
  \vspace{-4ex}
  \caption{\textbf{Fourier analysis of noisy latents under canny edge and depth map conditions}. (\textbf{\textit{Left}}) Average L2 distance between natural and condition image DFT spectra over timesteps.  Subfigures (a)–(d) and (e)–(h) show the DFT spectra of four randomly selected images for both conditions at different timesteps. In each group, (a/e) and (b/f) correspond to condition latents at $t_{\text{low}}$ and $t_{\text{high}}$, while (c/g) and (d/h) correspond to natural latents at $t_{\text{low}}$ and $t_{\text{high}}$, respectively. (\textbf{\textit{Right}}) Average high-frequency component ratio over timesteps.}
  \label{fig:canny_and_depth_dft}
  \vspace{-1.5em}
\end{figure*}

\section{Additional Related Work} \label{subsec:tbcontrollable}
Training-based controllable diffusion models train auxiliary modules or fine-tune the model to incorporate additional input signals to guide the generation process. Similar to \cref{subsec:tfcontrollable} in the main paper, we categorize them into three groups: \textbf{(i) Image editing}~\citep{brooks2023instructpix2pix, goel2023pairdiffusion, kim2022diffusionclip, wang2023imageneditor, sheynin2024emuedit, geng2024instructdiffusion, Xiao2025Omnigen, wu2025omnigen2, chen2025unireal, Le2025Onediffusion, Xia2025Dreamomni, xie2025anyrefill, han2024ace}; \textbf{(ii) Image-to-image translation}~\citep{isola2017im2im, saharia2022palette, tumanyan2022splicing, ouyang2025klora, park2019spade}; \textbf{(iii) Conditional text-to-image (T2I) generation} methods synthesize images that satisfy both a text prompt and a control condition. Among these approaches, some works condition the generation on layout cues (\eg, bounding boxes)~\citep{li2023gligen, yang2023reco, wang2024instancediffusion} or reference images of specific subjects~\citep{gal2022ti, ruiz2023dreambooth, ruiz2023hyperdreambooth, avrahami2023break-a-scene, po2023orthogonal, li2023blipdiffusion, Zhang2024SSRencoder, zhang2025easycontrol, tan2025ominicontrol, tan2025ominicontrol2, chen2025unireal, Xia2025Dreamomni, Xiao2025Omnigen, wu2025omnigen2, Le2025Onediffusion}. Another line of work~\citep{zhang2023controlNet, mou2023t2i, ye2023ipadapter, zhao2023unicontrol, avrahami2023spatext, Zhang2024SSRencoder, zhang2025easycontrol, tan2025ominicontrol, tan2025ominicontrol2, li2024controlnet_plus_plus, Xiao2025Omnigen, wu2025omnigen2, chen2025unireal, Le2025Onediffusion, Xia2025Dreamomni, xie2025anyrefill, han2024ace, xu2025ctrlora, peng2025local} enables fine-grained structural control by leveraging condition images of different modalities (\eg, canny edges, OpenPose keypoints~\citep{cao2019openpose}). Despite their impressive performance, these methods all require retraining or fine-tuning on datasets tailored to the control signal, which limits their generalization to new model checkpoints and novel control conditions.

\section{Additional Analyses}
\label{supp:sec:analysis}
\subsection{KL Divergence}
\label{supp:subsec:kldiv}
To analyze the domain gap between natural images and condition images, we collect 20 natural images from the \textit{ImageNet-T2IR} dataset from~\citep{tumanyan2023plug-and-play}. Then we use the ControlNet processor~\citep{zhang2023controlNet} to convert these natural images into 5 conditions (canny edge, depth map, normal map, HED edge, and scribble drawing), resulting in 100 natural-condition image pairs.

To quantify the distributional difference, we extract diffusion features at a fixed timestep for each image, flatten them into feature maps (size $(HW) \times F$), and concatenate all features from each domain. We then apply PCA to the combined feature set and retain only the first principal component. Each image is thus projected into a 1-dimensional vector of $HW$ values along this dominant component.

We estimate a probability distribution over these projections for each domain using Gaussian KDE. Specifically, we sample 1000 evenly spaced points between the minimum and maximum values observed in the two distributions. We then compute the KL divergence between the estimated densities of condition and natural images:
\begin{equation}
\mathrm{KL}(P \| Q) = \sum_{i=1}^{1000} p(x_i) \log \frac{p(x_i)}{q(x_i)} ,
\end{equation}
where $p(x)$ and $q(x)$ denote the normalized KDE densities of condition and natural images, respectively. We repeat this computation across timesteps to observe how the domain gap evolves during the diffusion process.

\subsection{Self-Similarity}
Following~\citep{tumanyan2022splicing}, we adopt the DINO self-similarity distance~\cite{caron2021dino} to quantify structural similarity between images. 
In Vision Transformer (ViT)~\citep{ViT}, an image is first divided into a sequence of non-overlapping patches, which are then linearly embedded and processed as tokens. In each Transformer layer, the tokens are projected into queries, keys, and values as follows:
\begin{equation}
    \mathbf{Q}_l = \mathbf{T}_{l-1}\mathbf{W}_{l}^Q,\; \mathbf{K}_l = \mathbf{T}_{l-1}\mathbf{W}_{l}^K ,\; \mathbf{V}_l = \mathbf{T}_{l-1}\mathbf{W}_{l}^V,
\end{equation}
where $\mathbf{T}_{l}(\image)$ denotes the output tokens for layer $l$ for image $\image$, and $\mathbf{W}_{l}^Q$, $\mathbf{W}_{l}^K$, and $\mathbf{W}_{l}^V$ are the corresponding query, key, and value weight matrices, respectively.

To capture an image’s internal structure, we compute its DINO self-similarity matrix at the final Transformer layer $L$:
\begin{equation}
S_L(\image)_{ij} = \mathrm{cos\_sim}\left( k_{L}(\image)_i, k_{L}(\image)_j \right),
\label{eq:selfsim}
\end{equation}
where $\mathbf{K}_{L}(\image)\!=\!\left[k_{L}(\image)_{\textit{cls}}, k_L(\image)_1,\dots, k_{L}(\image)_n\right]$ are the key embeddings from the last layer for image $\image$ ($n$ denotes the number of patch tokens), and $\mathrm{cos\_sim}$ denotes cosine similarity.

As shown in~\citep{tumanyan2022splicing}, this self-similarity-based descriptor can effectively capture the structure of an image while ignoring appearance details. Given two images $\image_1$ and $\image_2$, their structural distance is computed as the $\ell_2$ distance between their self-similarity matrices:
\begin{equation} \label{eq:5}
    \mathcal{L}^{\text{struct}} = \left\|S_{L}(\image_1) - S_{L}(\image_2) \right\|_2,
\end{equation}
where $S_{L}(\image)$ is defined in Eq.~(\ref{eq:selfsim}).

\subsection{Discrete Fourier Transformation (DFT)}
\label{supp:subsec:dft_analysis}

As an alternative to quantifying the domain gap between natural and condition images, we employ the Discrete Fourier Transformation (DFT) to analyze differences in their frequency components. Specifically, we begin by extracting diffusion feature maps for natural and condition images at a fixed timestep, following the method described in \cref{supp:subsec:kldiv}. Since DFT typically operates on spatial images rather than high-dimensional feature tensors, we use the diffusion decoder to transform these feature maps back into RGB images. We then apply DFT to the decoded images to obtain their frequency spectra and compute the L2 distance between the spectra of natural and condition image pairs.

\begin{figure}[h]
\centering
\includegraphics[width=\linewidth]{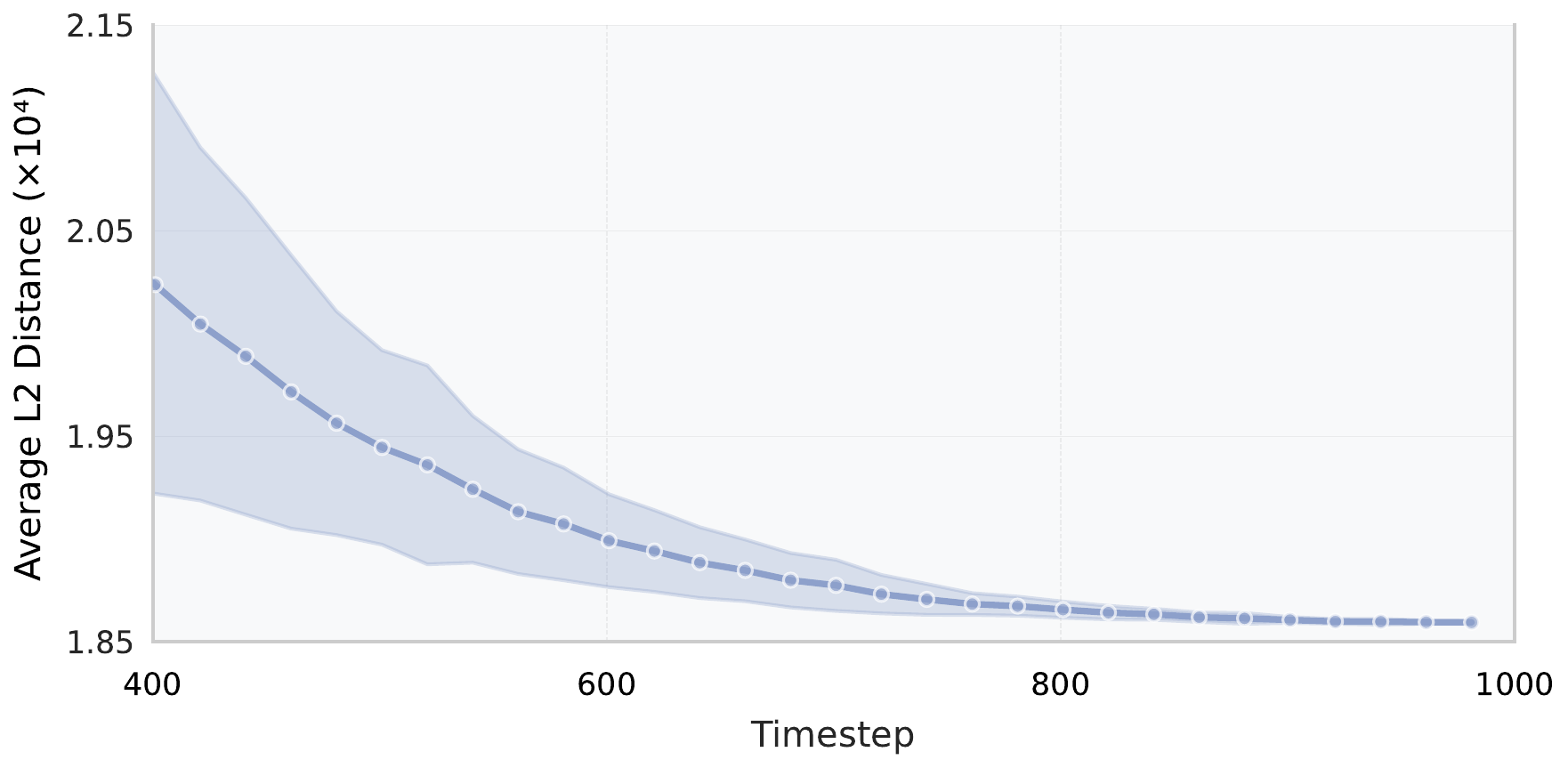}
\caption{\textbf{Average L2 distance between natural and condition image DFT spectra over diffusion timesteps.} Results are averaged over all five conditions.}
\label{fig:average_dft_distance}
\vspace{-1em}
\end{figure}

This process is repeated across all diffusion timesteps, and the resulting distances are averaged over the same 100 natural-condition image pairs as described in \cref{supp:subsec:kldiv}. As shown in \cref{fig:average_dft_distance}, the average L2 distance between the frequency spectra decreases progressively as the diffusion timestep increases. This trend indicates that the diffusion process gradually reduces the frequency-domain gap between natural and condition images---consistent with our findings in \cref{sec:revisit} in the main paper.

To further investigate how frequency components evolve through the diffusion process, we conduct a detailed analysis on two representative conditions: canny edge and depth map. Intuitively, canny edges, characterized by sharp edges and detailed contours, are expected to exhibit a higher proportion of high-frequency components in their DFT spectrum. In contrast, depth maps tend to be dominated by smooth gradients, suggesting a stronger presence of low-frequency components. 

As illustrated in \cref{fig:canny_and_depth_dft}, the average L2 distance between the DFT spectra of natural and condition latents decreases over time for both conditions, consistent with the trend shown in \cref{fig:average_dft_distance}. We also visualize DFT spectra of both image types at two representative timesteps of the denoising trajectory, denoted as $t_{\text{low}}$ and $t_{\text{high}}$. In practice, we set $t_{\text{low}} = 1$ and $t_{\text{high}} = 981$. Since SDXL inference performs 50 denoising steps, steps 1 and 981 correspond to the lowest and highest noise levels in the denoising process. At $t_{\text{low}}$, canny edge spectra exhibit dispersed high-activation regions, indicative of prominent high-frequency composition. In contrast, depth map spectra show energy concentrated near the center, reflecting low-frequency dominance. Both differ markedly from the corresponding spectra of natural images. At $t_{\text{high}}$, due to accumulated noise, the DFT spectra for all images become visually similar. 

This pattern is further confirmed by the right panel of \cref{fig:canny_and_depth_dft}, which plots the ratio of high-frequency components---defined as the proportion of DFT energy outside a centered circle with a radius equal to one-sixth of the image size---over timesteps. Initially, canny features are dominated by high-frequency content, while depth exhibits more low-frequency patterns. These differences gradually converge, reflecting a narrowing frequency-domain gap between different conditions.

\begin{figure*}[t]
  \centering
  \includegraphics[width=\linewidth]{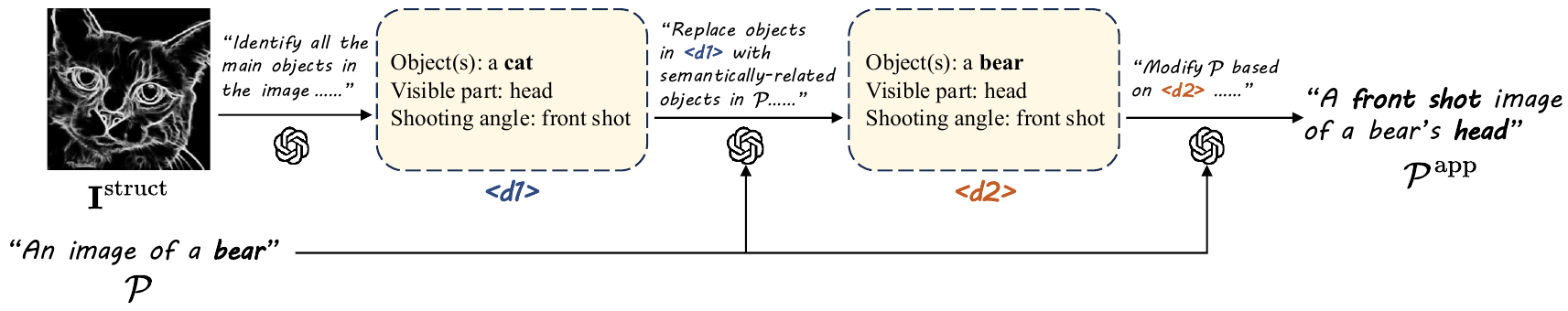}
  \caption{\textbf{Illustration of the Appearance-Rich Prompting (ARP) module}. Given the original text prompt $\prompt$, our module derives an appearance-rich prompt $\promptapp$ by integrating semantic information from the condition image $\imagecond$.}
  \label{fig:ARP_pipeline}
  \vspace{-1.5em}
\end{figure*}

\section{Method Details}
\label{supp:sec:method}
\subsection{Spatially-Aware Appearance Transfer}

We build our method on top of the spatially-aware appearance transfer mechanism proposed in Ctrl-X~\citep{lin2024ctrlx}. Specifically, given diffusion features $\mathbf{f}_{l,t}^\text{out}$ and $\mathbf{f}_{l,t}^\text{app}$ from the output and appearance branches respectively, Ctrl-X~\citep{lin2024ctrlx} computes a cross-image attention map as follows:
\begin{equation}
\begin{gathered}
    \mathbf{A} = \mathrm{softmax}\left( \frac{\mathbf{Q}^\text{out} {\mathbf{K}^\text{app}}^\top}{\sqrt{d}} \right), \\
    \mathbf{Q}^\text{out} = \mathrm{norm}(\mathbf{f}_{l,t}^\text{out}) \mathbf{W}_l^Q, \quad \mathbf{K}^\text{app} = \mathrm{norm}(\mathbf{f}_{l,t}^\text{app}) \mathbf{W}_l^K,
\end{gathered}
\end{equation}
where $\mathrm{norm}(\cdot)$ is applied across the spatial dimension ($HW$) and removes global statistics across spatial dimensions to isolate structural correspondence.

Subsequently, attention-weighted statistics are computed from the appearance features:
\begin{equation}
\begin{gathered}
    \mathbf{M} = \mathbf{A} \mathbf{f}_{l,t}^\text{app}, \\
    \mathbf{S} = \sqrt{ \mathbf{A} (\mathbf{f}_{l,t}^\text{app} \odot \mathbf{f}_{l,t}^\text{app}) - (\mathbf{M} \odot \mathbf{M}) },
\end{gathered}
\end{equation}
which are then used to modulate the output features:
{
\setlength{\abovedisplayskip}{4pt}
 \setlength{\belowdisplayskip}{4pt}
\begin{equation}
    \mathbf{f}_{l,t}^\text{out} \leftarrow \mathbf{S} \odot \mathbf{f}_{l,t}^\text{out} + \mathbf{M}.
\end{equation}
}

\subsection{Appearance-Rich Prompting}

Directly using the original prompt for appearance transfer may lead to artifacts in the generated image, since such prompts tend to be brief and lacking in semantic correspondence with the condition image (see \cref{sec:method_prompt} for details). To overcome this limitation, we propose a pipeline that enriches the original text prompt $\prompt$ with semantic information extracted from the structure condition image $\imagecond$, yielding a more appearance-rich prompt $\promptapp$ for generating the final appearance image $\imageapp$. As illustrated in \cref{fig:ARP_pipeline}, we first utilize GPT-4o~\cite{achiam2023gpt4} to extract key semantic entities from the condition image to produce dictionary $\langle d1\rangle$. To facilitate semantic alignment between the condition image and the text prompt, we further employ GPT-4o~\cite{achiam2023gpt4} to identify and associate these extracted entities with semantically-related elements in the original text, modifying $\langle d1\rangle$ to produce $\langle d2\rangle$. Finally, we revise the original prompt $\prompt$ using the extracted semantic information, producing an enhanced appearance prompt $\promptapp$. To help the multimodal LLM correctly follow instructions and mitigate erroneous semantic transfer, our pipeline stores intermediate information in structured dictionaries, enabling more controlled and interpretable prompt editing. More examples of the Appearance-Rich Prompting (ARP) module are provided in \cref{fig:supp_ablation_prompt}. For the full prompt used with GPT-4o~\cite{achiam2023gpt4} for Appearance-Rich Prompting, see the accompanying .txt file in the supplementary material.

\section{Experiment Setup Details}
\label{supp:sec:details}

\subsection{Implementation Details}
We implement our method with Diffusers~\citep{vonplaten2022diffusers} on SDXL 1.0~\citep{podell2023sdxl} and adopt the same injection layers following previous work~\citep{lin2024ctrlx}. We sample $\imagegen$ with 50 steps of DDIM sampling and set $\eta = 1$~\citep{song2020ddim}. For structure-rich injection, we perform injection when $t \ge \injlast$, and we set $\injlast=400$ and $C=600$. 

For restart refinement, we set $\restartstdmin=1.0$, $\restartstdmax=2.0$, $\restarttimes=3$, $\restartsteps=5$, where $\restartsteps$ is the total number of timesteps in the restart backward process. For the restart backward process, we adopt the same noise schedule as the base model, SDXL~\citep{podell2023sdxl}, which is:
\begin{align}
\small
\sigma_{\min} = \sqrt{\frac{\beta_{\min}}{1-\beta_{\min}}},
\quad
\sigma_{\max} = \sqrt{\frac{\beta_{\max}}{1-\beta_{\max}}},
\end{align}
\vspace{-0.5em}
\begin{align}
\small
\sigma_t = \sigma_{\min} - \left(\sigma_{\max}-\sigma_{\min}\right)\frac{t}{T-1}, \nonumber \\
\alpha_t = \frac{1}{1+\sigma_t^2},
\quad
\beta_t = 1 - \alpha_t,
\end{align}

where $\beta_{\min} = 0.00085$ and $\beta_{\max} = 0.012$. 

For self-recurrence, we set $\recurtmin=500$, $\recurtmax=900$, $\recurtimes=2$, where $\recurtmax$ is the self-recurrence starting point, $\recurtmin$ is the self-recurrence end point, and $\recurtimes$ is the number of self-recurrence~\citep{lin2024ctrlx}. We run most experiments on NVIDIA Tesla V100 GPUs. For FreeControl~\citep{mo2023freecontrol}, InfEdit~\citep{xu2023infedit}, and computational efficiency comparisons, we run the experiments on A800 GPUs.

For any input condition image $\imagecond$, we preprocess it with a dilation and unsharp masking operation. Specifically, we binarize the image, perform a distance transform operation to detect the minimum line width $\width$. If $\minwidth \leq \width \leq \maxwidth$, we dilate $\imagecond$ with kernel size $\kernelsize$. On the other hand, if the inverted image meets the standard, we erode $\imagecond$. We set $\minwidth=25$, $\maxwidth=50$ and $\kernelsize=10$. Then we perform unsharp masking $(1 + \gamma) \cdot \textbf{I}^{\text{dilate}} - \gamma \cdot B$ to modify the dilated (eroded) image, where $\textbf{I}^{\text{dilate}}$ denotes the dilated (eroded) input condition image, $\gamma=50$, and $B$ is the Gaussian blur operation with blur radius $r=3$. We empirically find the two operations beneficial for highlighting object boundaries and improving structure preservation. 

For implementation on FLUX~\citep{flux2024}, we use the FLUX.1-dev pipeline as the backbone. We use guidance strength $s=6.5$ and 25 inference steps. For SRI, we inject self-attention query matrices for structural control~\citep{lin2025freecontrol}, applied at timesteps $t \ge \injlast$ with $\injlast=600$. We found that a synchronous schedule $\asyncfunc=t$ performs slightly better than the constant schedule $\asyncfunc=C$, so we adopt the former in our experiments. Injection is performed at layers 25--38 of FLUX. For RR and ARP, we adopt the same settings as described above.

\subsection{Dataset Details}

We construct our dataset based on the conditional generation datasets from Ctrl-X~\citep{lin2024ctrlx} and FreeControl~\citep{mo2023freecontrol}. Specifically, for conditions canny edge, depth map, normal map, HED edge, and scribble drawing, we select condition-prompt pairs from both datasets and merge them. We collect a total of 15 condition images per condition and form 22 condition-prompt pairs for canny edge and 21 pairs for each of the remaining four conditions.

For human pose and segmentation map, since both original datasets contain limited examples, we supplement them by collecting additional human pose images from the web and segmentation masks from the ADE20K~\citep{Zhou_2017_CVPR} dataset. We obtain 15 images for each of these two conditions and pair them with text prompts using a combination of templates and hand annotation, resulting in 21 image-prompt pairs for human pose and 23 for segmentation mask.

\subsection{User Study Details}

\begin{figure*}[t]
  \centering
  \includegraphics[width=0.95\linewidth]{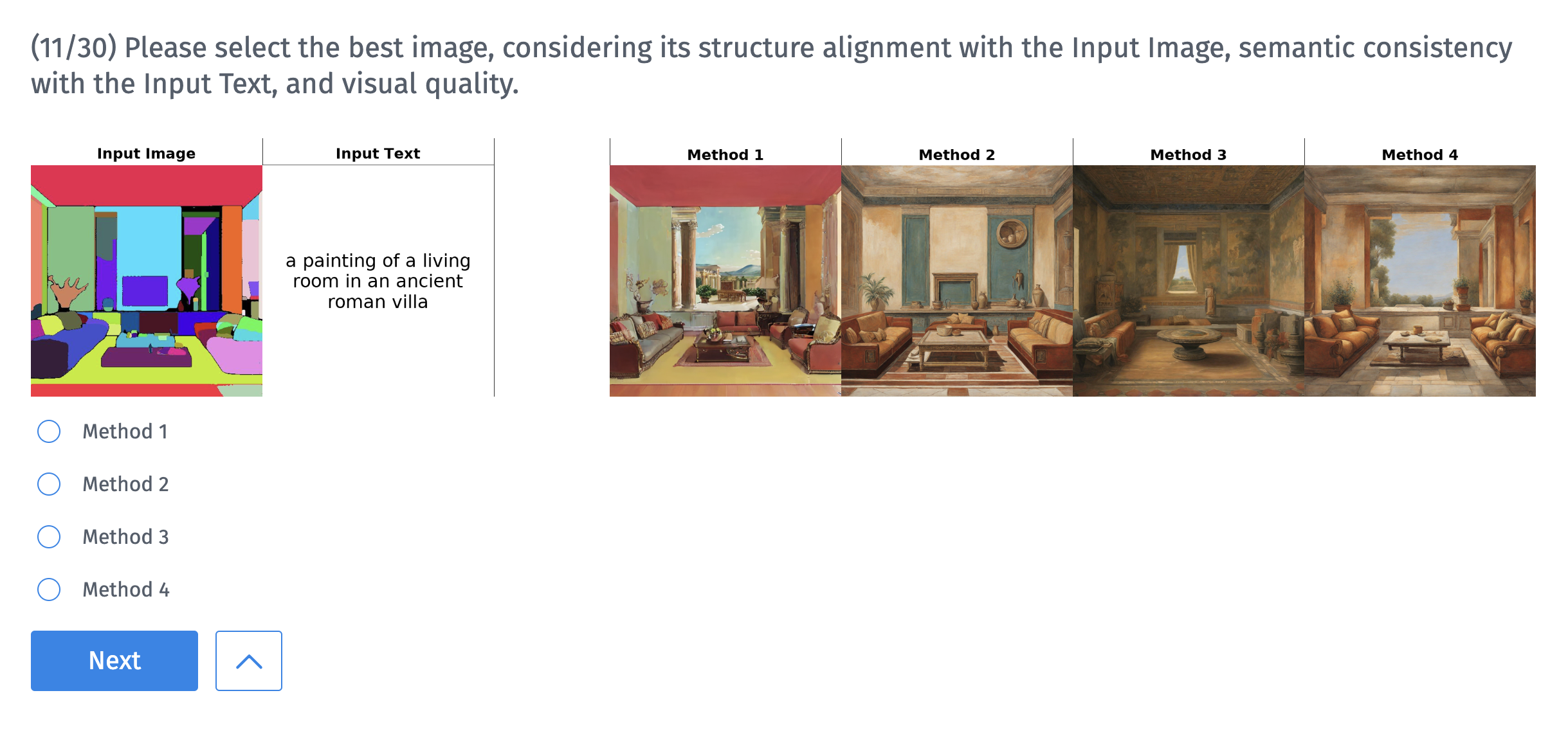}
  \vspace{-1ex}
  \caption{\textbf{Screenshot of the user study interface.} Participants are presented with the inputs and asked to select the best result from four randomly shuffled candidates.}
  \label{fig:user_study_interface}
\end{figure*}

To ensure fairness in the user study comparison, we randomly sample 30 cases from our dataset. For each case, 40 participants with relevant expertise are asked to select the most preferred image from four anonymous, randomly shuffled results according to a holistic criterion accounting for structure alignment with the condition image, semantic consistency with the prompt, and visual quality. The instruction provided in the questionnaire is: \textit{Please select the best image, considering its structural alignment with the Input Image, semantic consistency with the Input Text, and visual quality.} \cref{fig:user_study_interface} shows the interface of the user study.

\subsection{Computational Efficiency Experiment Details}

We evaluate the baselines on our dataset to compare their average inference time and memory usage. Specifically, we implement FreeControl~\citep{mo2023freecontrol}, Ctrl-X~\citep{lin2024ctrlx}, and our method using SDXL 1.0~\citep{podell2023sdxl} checkpoints. For InfEdit~\citep{xu2023infedit}, we utilize the LCM Dreamshaper v7~\citep{luo2023lcmdreamshaperv7} checkpoint (based on SD1.5), as it is the only model provided in their official codebase. To ensure a fair comparison, we generate $1024 \times 1024$ images using $50$ sampling steps for all methods.

\subsection{Evaluation Metric Details}

We adopt three traditional metrics for evaluation, each targeting a different aspect of generation quality. (i) CLIP score~\citep{radford2021clip} measures text-image alignment via cosine similarity between the CLIP embeddings of the generated image and text prompt. A higher score indicates a better adherence to the textual description. (ii) DINO self-similarity score (Self-sim)~\citep{tumanyan2022splicing} quantifies structural alignment by computing the distance between the generated image and the original RGB image used to derive the structural condition in DINO-ViT feature space. (iii) Condition LPIPS score (LPIPS)~\citep{zhang2018lpips} measures perceptual deviation between the generated image and the condition image. 
A higher score indicates greater deviation, reflecting less appearance leakage from the condition input and higher generation quality. 

\subsection{SRI Ablation Study Details}

To comprehensively evaluate the effect of the sampling schedule on controllable generation, we adopt a systematic design space spanning schedules of varying complexity. We begin with constant schedules $\asyncfunc = C$ for several values of $C$ as the simplest form, then evaluate linear functions with different slopes and intercepts. To cover more expressive functional forms, we further consider three representative families: (i) quadratic functions, representing concave behavior; (ii) logarithmic functions, representing convex behavior; and (iii) trigonometric functions, representing oscillatory behavior. Together, these schedules cover a broad range of monotonicity, curvature, and rate of change throughout the injection timestep range. As shown in \cref{fig:ablation_injection} in the main paper, regardless of the specific functional form, schedules whose values predominantly lie around medium timesteps consistently yield optimal performance, with the simplest such instance being the constant schedule $C=600$. In contrast, schedules with rapid variation, such as oscillatory ones, perform notably worse. These observations suggest that the overall distribution of its values matters more than the specific shape of the schedule.

\section{Additional Experimental Results}
\label{supp:sec:exp}

\subsection{Computational Efficiency}
\label{supp:subsec:inference}

Complementing the overall efficiency comparison against prior works in~\cref{tab:compute_and_user} in the main paper,  we further analyze the specific runtime contribution of each module within our pipeline in~\cref{tab:inference-time}. 
The SRI module, which dominates the computation (85.1\%), represents the core injection framework responsible for handling both condition and appearance image features. 
Notably, the additional latency introduced by the RR and ARP modules is marginal, accounting for only 6.8\% and 8.1\% of the total inference time, respectively. 
Despite their low computational overhead, these components play a critical role in significantly enhancing image quality, as evidenced by the ablation study in \cref{tab:ARP_RR_ablation} in the main paper.

\begin{table}[h]
\centering
\caption{Proportion of inference time consumed by each module of our method.}
\begin{tabular}{l c}
\toprule
Module & Percentage of Inference Time \\
\midrule
SRI & 85.1\% \\
RR & 6.8\% \\
ARP & 8.1\% \\
\bottomrule
\end{tabular}
\label{tab:inference-time}
\vspace{-1em}
\end{table}

\subsection{Additional Qualitative Results}

\begin{figure*}[h]
    \centering
    \includegraphics[width=0.8\linewidth]{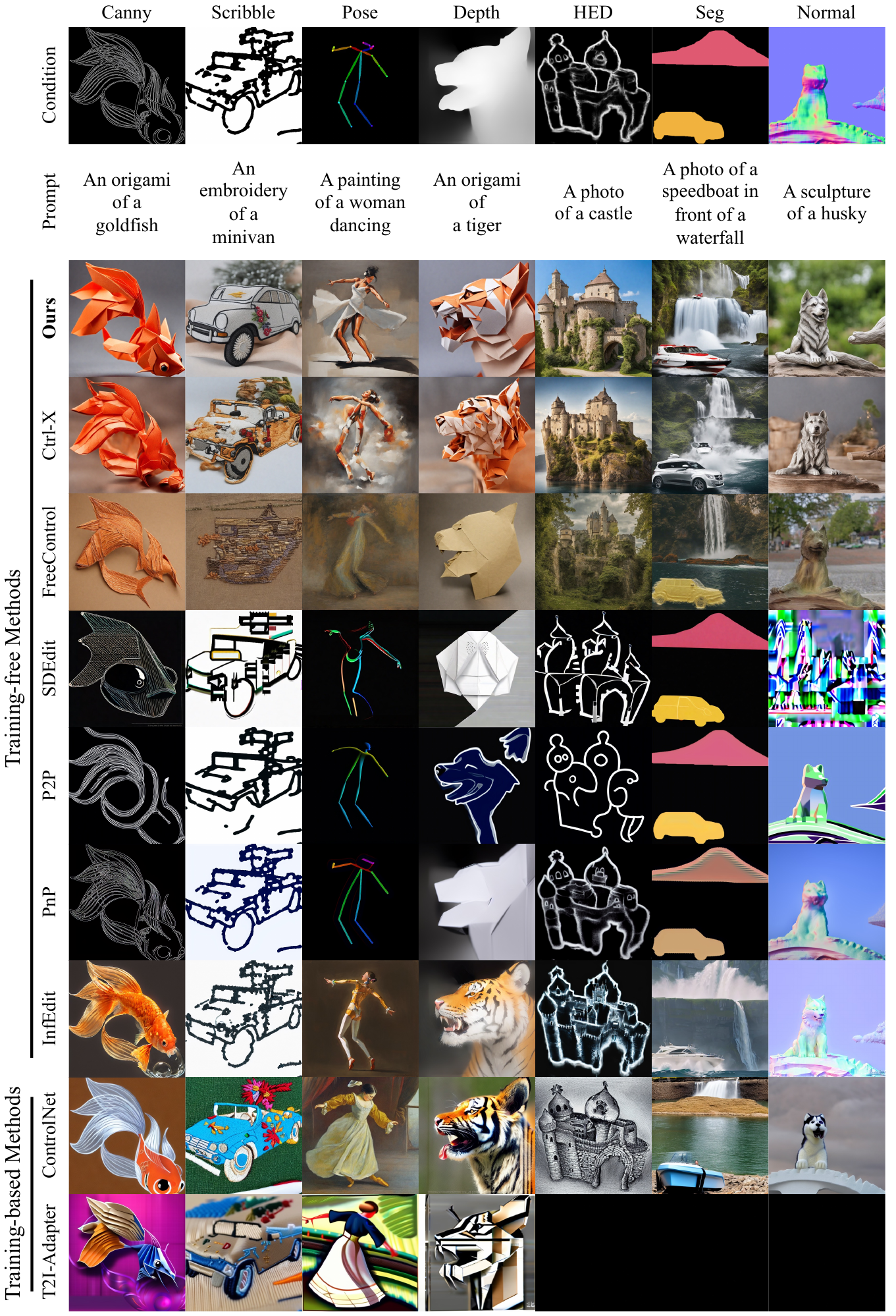}
    \caption{\textbf{Qualitative comparison with existing methods.}
    }
    \label{fig:compare_sota_more}
    \vspace{1cm}
\end{figure*}

\begin{figure*}[h]
    \centering
    \includegraphics[width=0.95\linewidth]{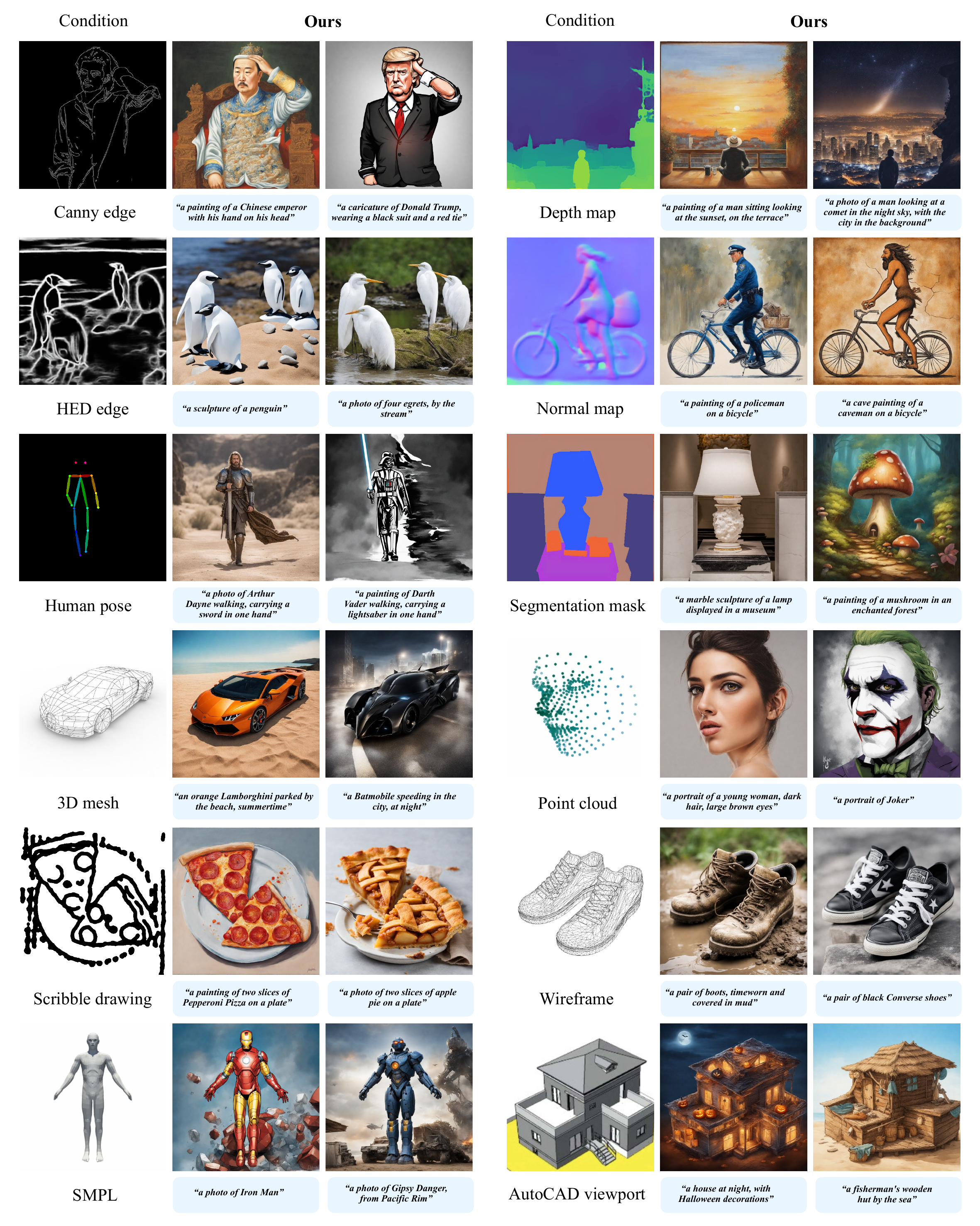}
    \vspace{-1ex}
    \caption{\textbf{Qualitative results for more control conditions.}}
    \label{fig:quality_result}
\end{figure*}

We provide additional qualitative comparisons with baselines in \cref{fig:compare_sota_more} and additional qualitative results for a broader range of condition types in \cref{fig:quality_result}. Our method demonstrates strong generation performance across both common and challenging conditions. It also handles diverse and complex text prompts effectively. As a training-free approach, it generalizes effortlessly to various in-the-wild conditions without any additional training cost, producing high-quality outputs. This level of zero-shot generalization is often unattainable for training-based methods.

\subsection{Additional Quantitative Results}
\label{supp:subsec:quantitative}
Since T2I-Adapter-SDXL~\citep{mou2023t2i} supports only four (canny, depth, normal, and pose) out of the seven condition types in our dataset, we further conduct a quantitative comparison limited to these four types. As shown in \cref{tab:sota}, our method outperforms all baselines across almost every metric. Notably, these metrics jointly assess both structure preservation (e.g., DINO self-similarity~\citep{tumanyan2022splicing}, DreamSim~\citep{fu2023dreamsim}) and generation quality (e.g., ImageReward~\citep{xu2023imagereward}, HPSv2~\citep{wu2023hpsv2}), highlighting the effectiveness of our approach.

\begin{table*}[t]
\centering
\caption{\textbf{Additional quantitative comparison of controllable T2I.} 
Our method consistently surpasses all training-free baselines in structure preservation, image-text alignment, and visual diversity. 
The best results are in \textbf{bold}, and the second best are \underline{underlined}.}
\small
\label{tab:sota}
\begin{tabular}{l c c c c c c}
\toprule
Method & Self-Sim $\downarrow$ & CLIP $\uparrow$ & LPIPS $\uparrow$ & \makecell[c]{Dream-\\Sim $\downarrow$} & \makecell[c]{Image-\\Reward $\uparrow$} & HPSv2 $\uparrow$ \\
\midrule
\graycell{ControlNet \citep{zhang2023controlNet}} & \graycell{0.067} & \graycell{0.309} & \graycell{0.701} & \graycell{0.509} & \graycell{0.298} & \graycell{0.285} \\
\graycell{T2I-Adapter \citep{mou2023t2i}}        & \graycell{0.116} & \graycell{0.287} & \graycell{0.728} & \graycell{0.636} & \graycell{-0.050} & \graycell{0.261} \\
\midrule
SDEdit \citep{meng2021sdedit} & 0.154 & 0.259 & 0.315 & 0.734 & -1.374 & 0.189 \\
P2P \citep{hertz2022prompt-to-prompt} & 0.197 & 0.251 & 0.266 & 0.724 & -1.786 & 0.168 \\
PnP \citep{tumanyan2023plug-and-play} & 0.157 & 0.256 & 0.151 & 0.724 & -1.789 & 0.168 \\
InfEdit \citep{xu2023infedit}         & 0.135 & 0.296 & 0.357 & 0.636 & -0.202 & 0.244 \\
FreeControl \citep{mo2023freecontrol} & 0.116 & \underline{0.320} & \textbf{0.667} & 0.626 & \underline{0.554} & \underline{0.285} \\
Ctrl-X \citep{lin2024ctrlx}           & \underline{0.104} & 0.315 & 0.650 & \underline{0.579} & 0.291 & 0.283 \\
\textbf{Ours}                         & \textbf{0.096} & \textbf{0.322} & \underline{0.662} & \textbf{0.558} & \textbf{0.897} & \textbf{0.313} \\
\bottomrule
\end{tabular}
\end{table*}

\subsection{Additional Ablation Study}
\begin{figure*}[htbp]
  \centering
  \includegraphics[width=0.85\linewidth]{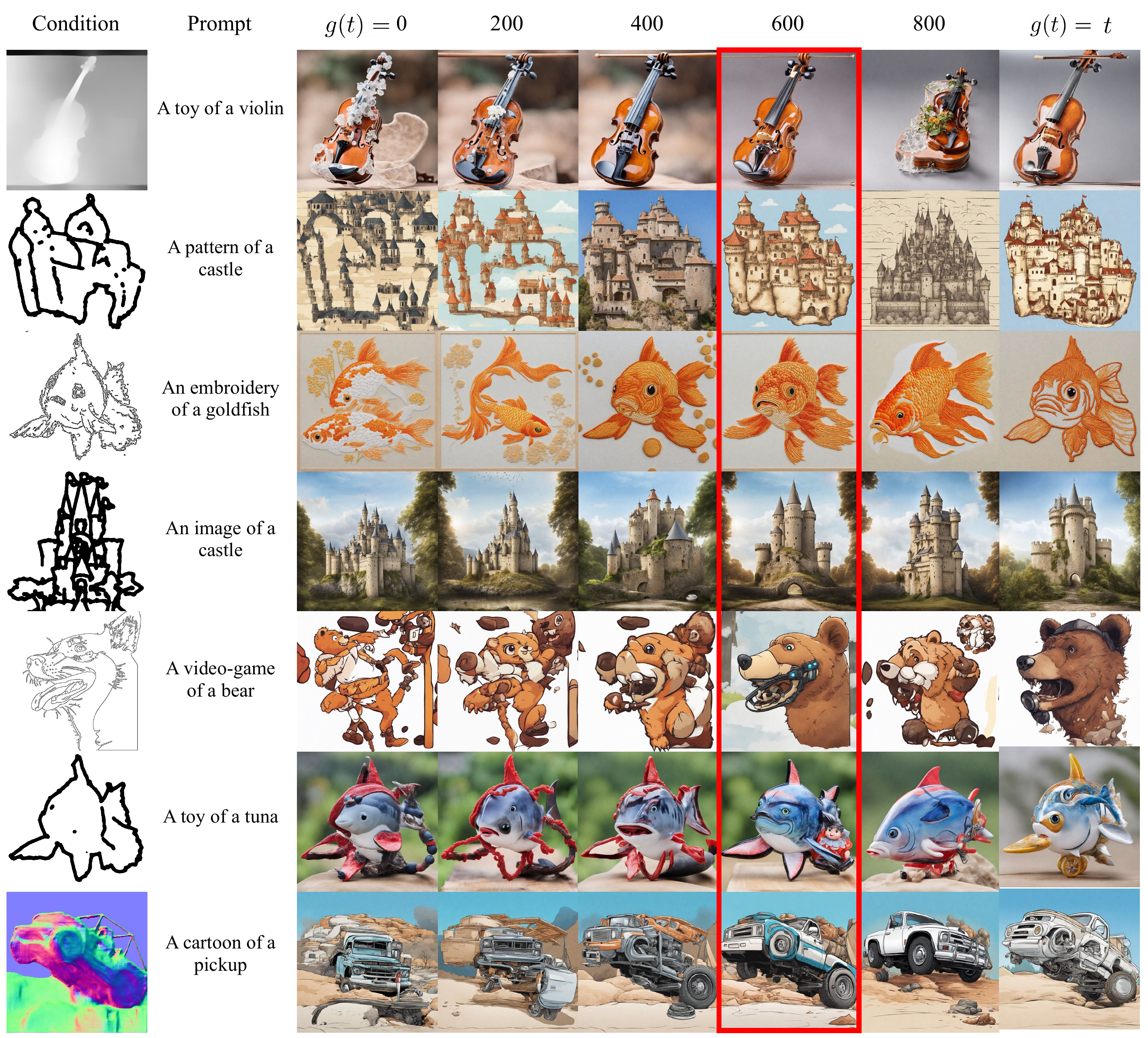}
  \vspace{2ex}
  \caption{\textbf{Additional ablation of Structure-Rich Injection (SRI)}. For asynchronous injection $\asyncfunc=C$, lower $C$ suffers from conditional leakage, while higher values improve appearance fidelity at the cost of structural control. The optimal trade-off is achieved at $C=600$, outperforming the synchronous schedule ($\asyncfunc=t$).}
  \label{fig:supp_ablation_injection}
\end{figure*}

\begin{figure*}[htbp]
  \centering
  \includegraphics[width=0.88\linewidth]{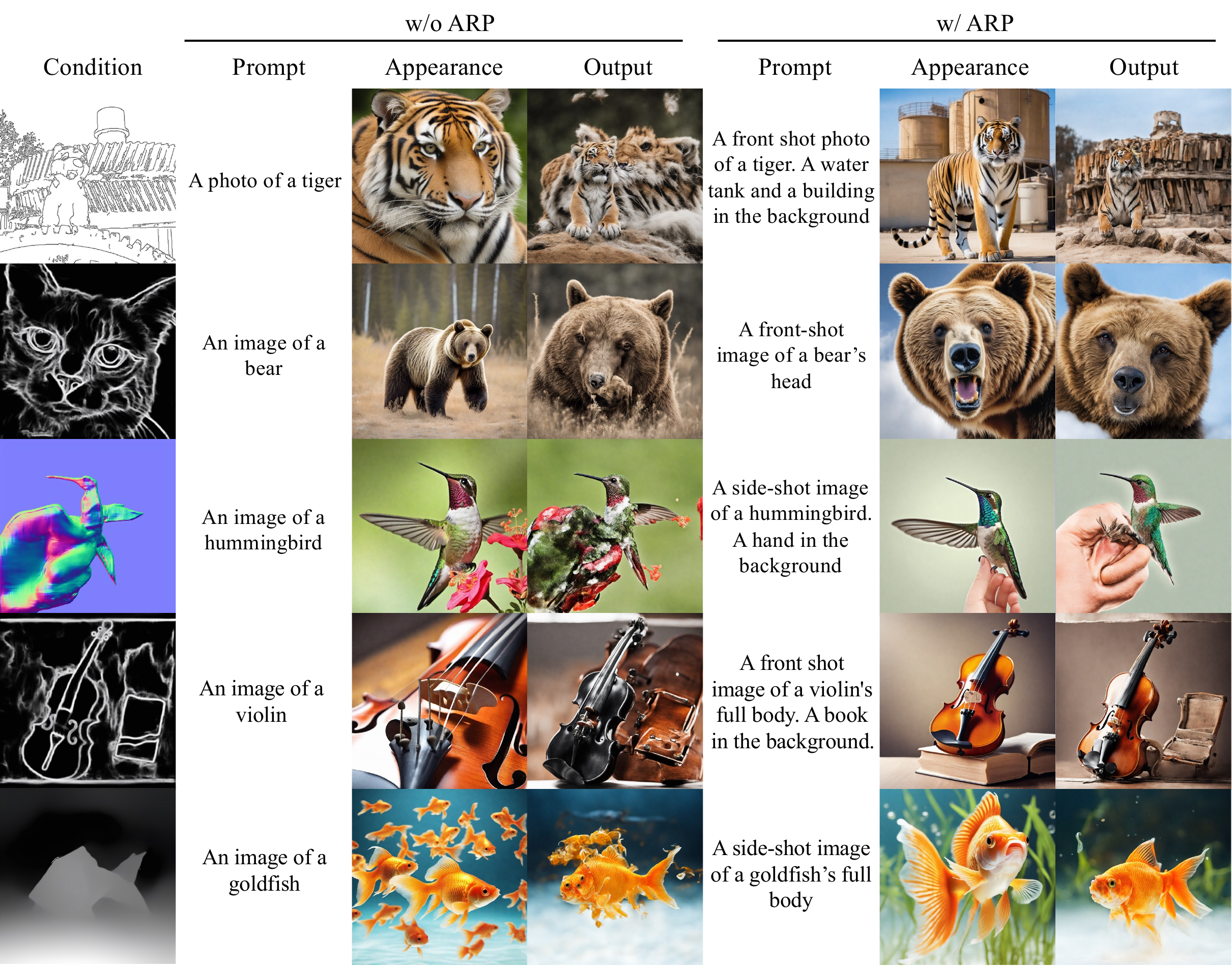}
  \vspace{1ex}
  \caption{\textbf{Additional ablation of Appearance-Rich Prompting (ARP)}. This module improves semantic alignment with the condition image by adapting prompts to better capture key visual attributes, thereby mitigating incorrect appearance transfers and reducing artifacts.}
  \label{fig:supp_ablation_prompt}
\end{figure*}

\begin{figure*}[htbp]
  \centering
  \includegraphics[width=0.88\linewidth]{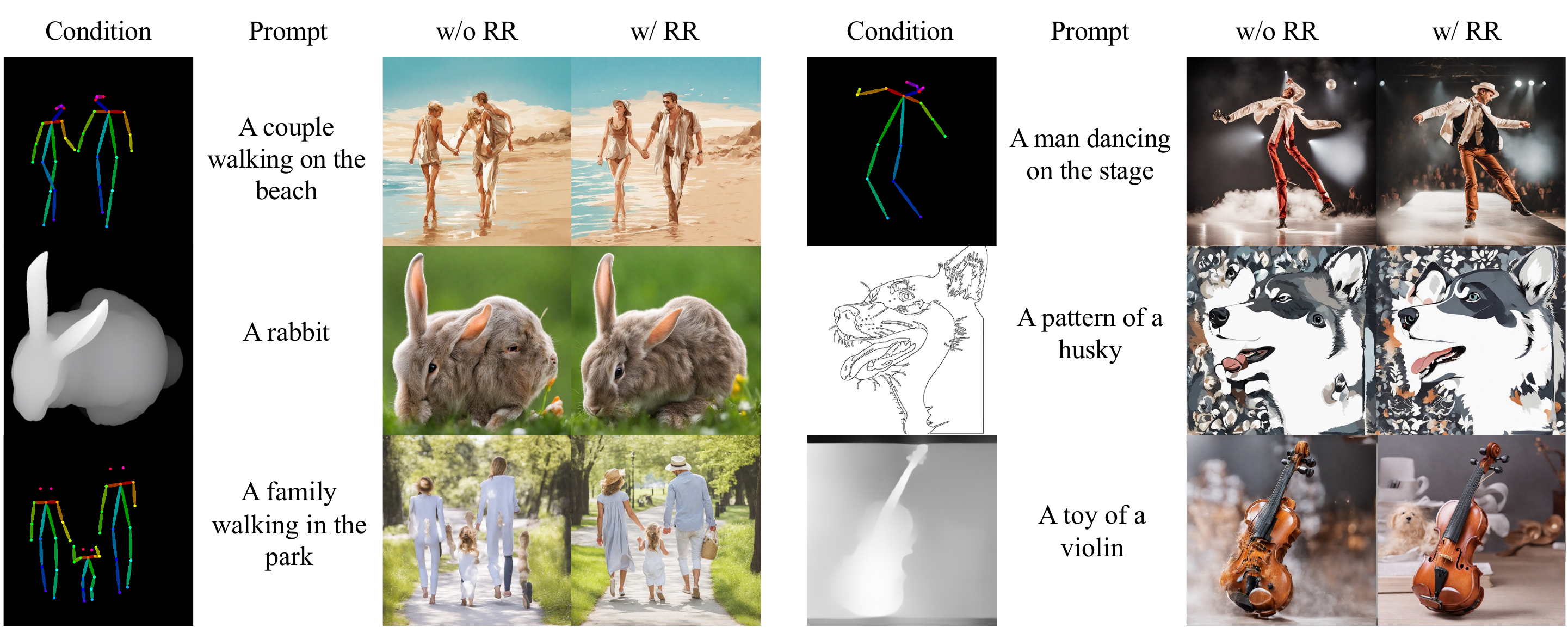}
  \vspace{1ex}
  \caption{\textbf{Additional ablation of Restart Refinement (RR)}. This strategy significantly mitigates condition leakage and appearance artifacts, improving generation quality while maintaining structural alignment.}
  \label{fig:supp_ablation_restart}
\end{figure*}

We present additional ablation studies on key components of our proposed method to validate our design choices. The results are shown in \cref{fig:supp_ablation_injection}, \cref{fig:supp_ablation_prompt}, and \cref{fig:supp_ablation_restart}. 

\vspace{0.5em}
\noindent\textbf{Structure-Rich Injection.}  
As a complementary study to the SRI ablation presented in \cref{sec:SRI_ablation} in the main paper, we further investigate the choice of constants in the case of constant injection. Specifically, we evaluate the effects of the injection schedule $\asyncfunc=C$ across various $C$ values. 
As shown in \cref{fig:supp_ablation_injection}, lower $C$ values result in severe conditional leakage due to a pronounced domain gap (\eg $C = 0$).  
In contrast, higher values of $C$ (\eg $C = 800$) produce more natural appearances with higher fidelity but compromise structural control.
Empirically, $C = 600$ achieves the best balance between appearance fidelity and structure control, significantly outperforming the synchronous injection baseline by enhancing structural alignment, suppressing condition leakage and reducing visual artifacts simultaneously.

\vspace{0.5em}
\noindent\textbf{Appearance-Rich Prompting.}
\cref{fig:supp_ablation_prompt} demonstrates the effectiveness of appearance-rich prompting in enhancing semantic alignment between the structural condition and the appearance image.
This strategy helps recover missing semantic elements (\eg, ``building'' in row 1 and ``hand'' in row 3), significantly reducing visual artifacts and improving the overall quality of the generated images.

\vspace{0.5em}
\noindent\textbf{Restart Refinement.} 
\cref{fig:supp_ablation_restart} illustrates the efficacy of restart refinement in mitigating visual artifacts (\eg,  the duplicated eye on the rabbit’s body and the incorrect eyes in the husky’s background). Additionally, it alleviates condition leakage under abstract conditions (\eg, pose), further improving generation fidelity.

\begin{figure}[htbp]
  \centering
  \includegraphics[width=0.95\linewidth]{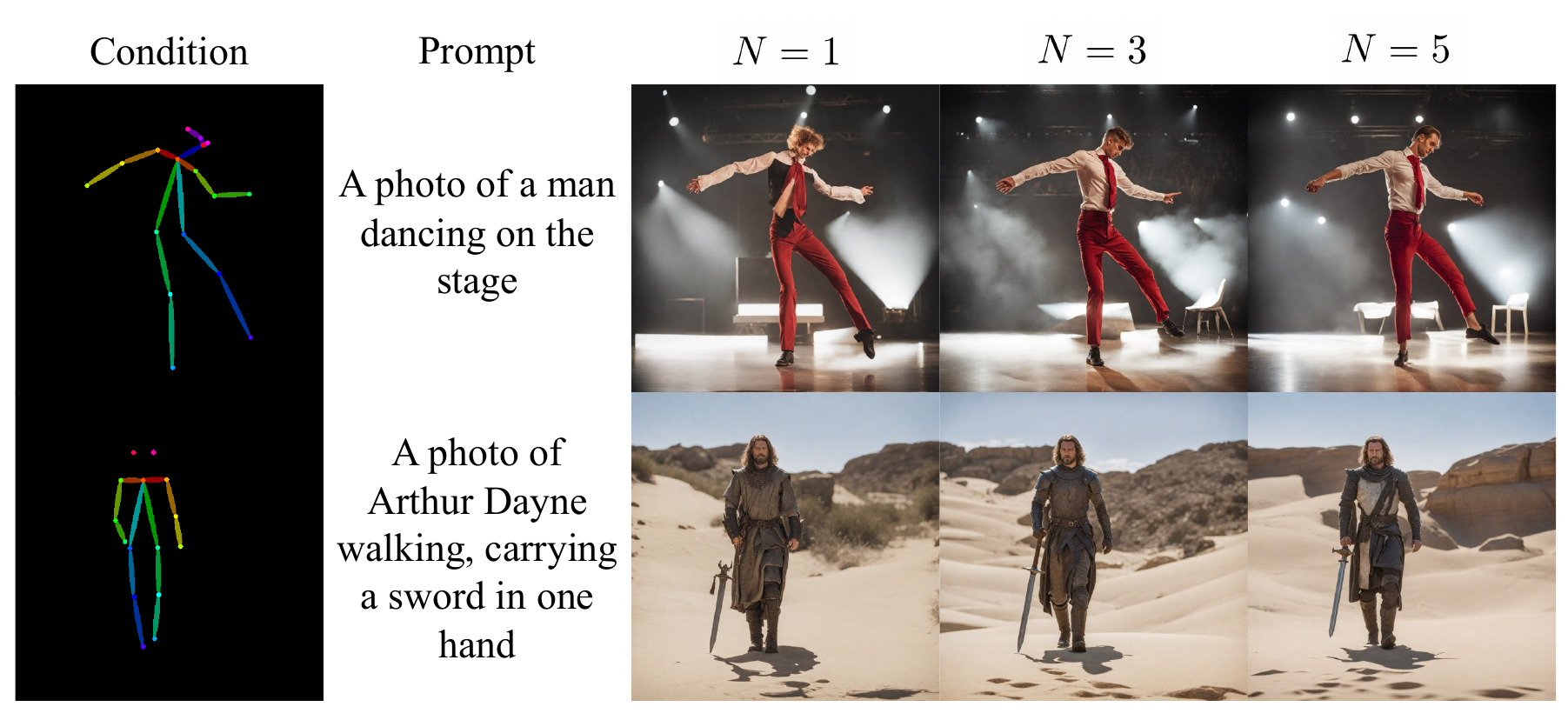}
  \caption{\textbf{Additional ablation of restart iterations $N$}. Setting $N=1$ is not adequate for suppressing visual artifacts, and both $N=3$ and $N=5$ yield high-quality outputs. Consequently, we set $N=3$ for optimal visual quality and computational efficiency.}
  \label{fig:supp_ablation_restart_N}
\end{figure}

\vspace{0.5em}
\noindent\textbf{Number of restart iterations $N$.}  We also conduct an ablation study of restart iterations $N$. As shown in \cref{fig:supp_ablation_restart_N}, setting $N=1$ is not adequate for suppressing visual artifacts, and both $N=3$ and $N=5$ yield high-quality outputs. Consequently, we set $N=3$ for optimal visual quality and computational efficiency.

\subsection{Additional Plug-in Results}

\begin{figure}[h]
  \centering
  \includegraphics[width=0.95\linewidth]{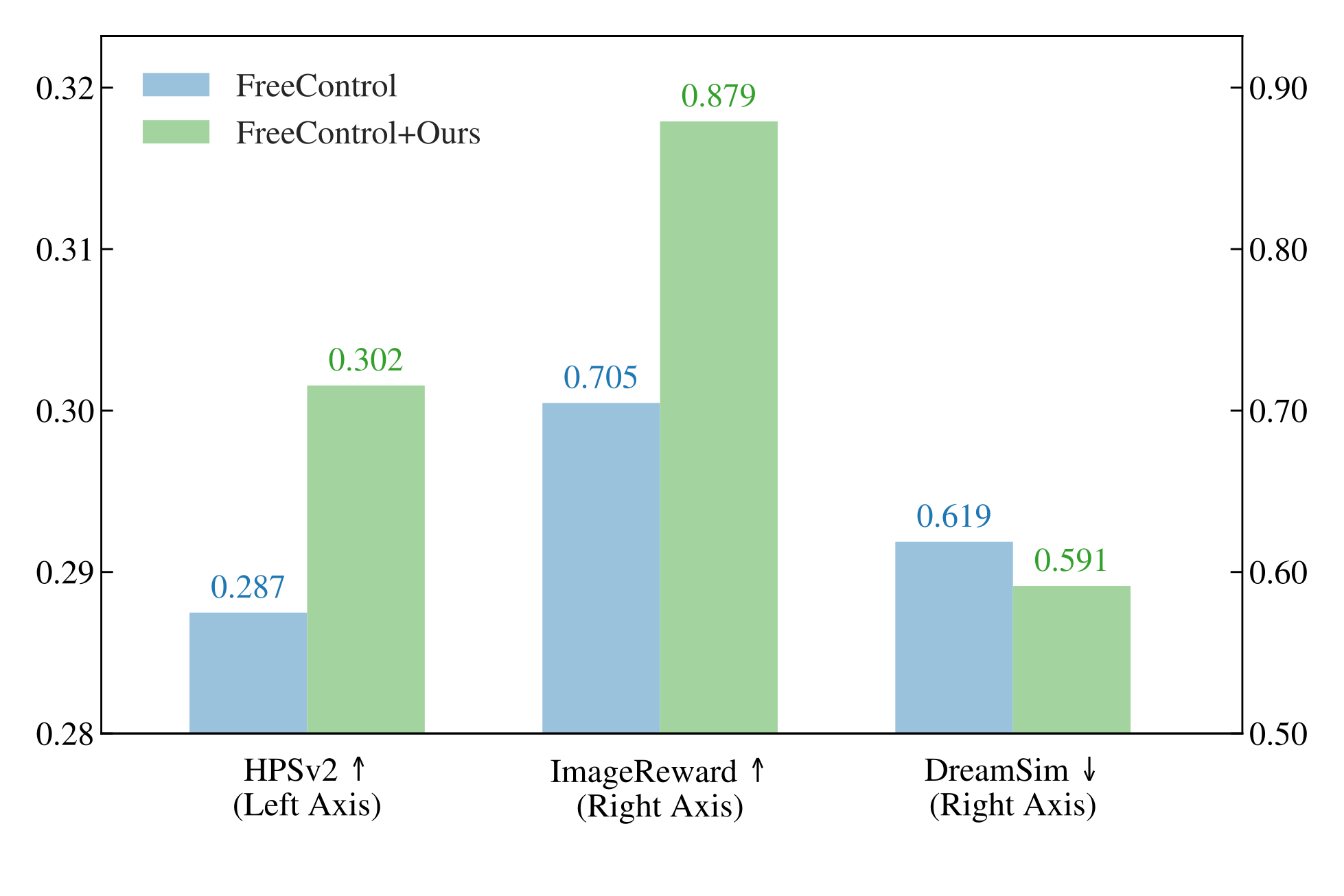}
  \caption{\textbf{Quantitative comparison of FreeControl~\citep{mo2023freecontrol} with and without our method}. Our approach consistently improves FreeControl on all three metrics.}
  \label{fig:supp_freecontrol_compare}
\end{figure}

We also compare FreeControl~\citep{mo2023freecontrol} with and without our method across three reward model metrics. As shown in \cref{fig:supp_freecontrol_compare}, our approach consistently improves FreeControl on all three metrics, demonstrating better structure alignment and appearance quality, further highlighting the flexibility of our framework.

\section{Broader Impacts} \label{supp:sec:broader}

Our method enables higher-quality and more realistic controllable generation without additional training or optimization. However, this capability and accessibility also raise the risk of malicious use of pretrained generative models (e.g., deepfakes). We urge the community not to misuse our method for deceptive or harmful purposes, such as spreading misinformation or generating non-consensual content. In response to such safety concerns, large generative models have increasingly incorporated safeguards. Likewise, our framework can inherit these protections, as it is built on a pretrained backbone, and its plug-and-play nature allows the open-source community to scrutinize and enhance its safety.

\clearpage

\end{document}